\documentclass[11pt]{article}

\usepackage[final]{acl}

\usepackage{times}
\usepackage{latexsym}
\usepackage{amsmath}
\usepackage{array}
\usepackage{booktabs}

\usepackage[T1]{fontenc}

\usepackage[utf8]{inputenc}

\usepackage{microtype}

\usepackage{inconsolata}

\usepackage{graphicx}

\title{From Apparent Search Gains to Selection Failures: Auditing Test-Time Inference in Structured Workflow Planning}

\author{
\textbf{Zhitong Chen}\textsuperscript{†}\quad
\textbf{Weifeng Zhang}\quad
\textbf{Kai Yin}\quad
\textbf{Zhiyuan Wang}\\
\textbf{Ali Mostafavi}\quad
\textbf{James Caverlee}\\
Texas A\&M University\\
\texttt{\{zhitong.chen18,weifengzhang,kai.yin,zhiyuan\}@tamu.edu}\\
\texttt{\{mostafavi,caverlee\}@tamu.edu}
}

\begin{document}
\maketitle
\begingroup
\renewcommand\thefootnote{\fnsymbol{footnote}}
\footnotetext[1]{Corresponding author.}
\endgroup
\begin{abstract}
Test-time search can improve structured planning, but when methods differ in prompts, post-processing, and selection, it is unclear what an observed gain should be attributed to. We study this question on DisasterBench, where models compose typed multi-tool workflows, by decomposing inference into three observable stages: whether a correct workflow is discovered, whether it survives structural validation, and whether it is finally selected. Comparing independent sampling, a nominal tree-search
configuration, and multi-operator search under controlled
inference budgets on two model families, we find that (i) the nominal tree-search configuration realizes
non-adaptive complete-workflow sampling, and independent
sampling with its expansion prompt reproduces its
candidate discovery without constructing a tree; (ii) structural validation never removes the last correct
workflow in our evaluated candidate pools, whereas
majority voting leaves a discovered correct workflow
unselected on 10--19\% of tasks per pool, with an audit
confirming substantive tool-choice or information-flow
errors in 57 of 60 examined cases; and (iii) a tool-specification-aware verifier recovers some of these losses but overturns more correct decisions than it fixes. Our results motivate auditing realized rather than nominal inference pipelines, and reporting discovery, validity, and selection separately.
\end{abstract}

\begin{figure*}[t]
    \centering
    \includegraphics[width=\textwidth]{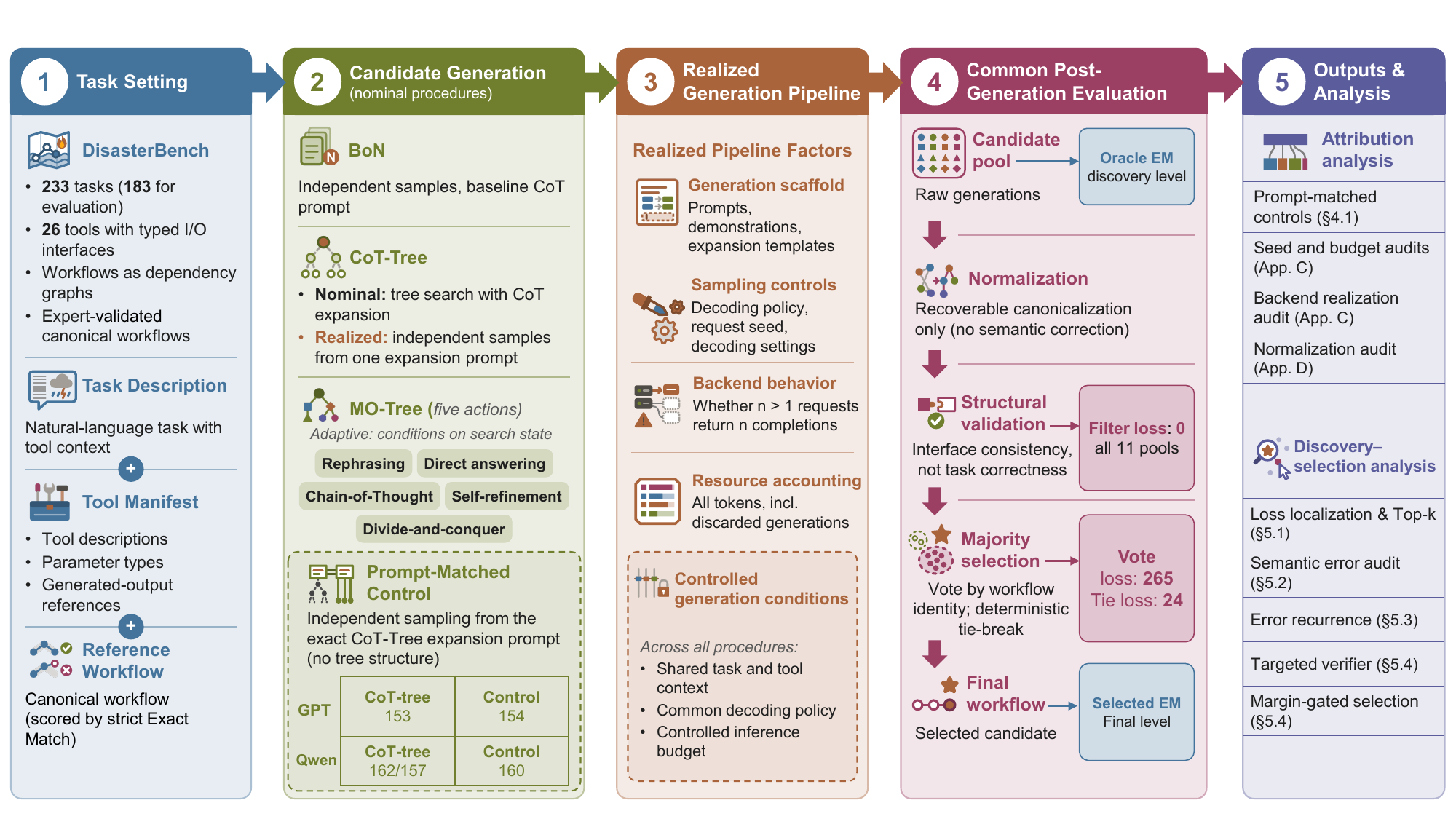}
    \caption{
    Overview of the experimental pipeline.
    Candidate-generation procedures are analyzed through their realized
    generation conditions and a common post-generation evaluation.
    Oracle EM measures candidate discovery, structural validation checks
    interface consistency, and majority selection produces the final
    workflow; the DVS decomposition localizes post-discovery failures
    as filter, vote, or tie losses.
    }
    \label{fig:overview}
\end{figure*}

\section{Introduction}

Disaster response increasingly draws on specialized AI capabilities
such as flood mapping, remote-sensing analysis, and damage assessment
\citep{gupta2019creating,bonafilia2020sen1floods11}, and language models are
now used to compose them into multi-step workflows by selecting tools
and routing information between them
\citep{chen2026disasterbench,cui2026large,wang2026maestro}. A plausible
workflow is not necessarily a correct one: a plan may feed an
object-segmentation map, rather than the generated image, into a
landslide-segmentation step, satisfying every interface constraint while routing
information incorrectly.

A natural way to improve reliability is to spend more computation at
inference time, through repeated sampling, structured search, or
downstream verification
\citep{wang2022self,yao2023tree,hao2023reasoning,cobbe2021training,snell2024scaling}.
Yet when such a method raises accuracy, two questions are usually left
open: where did the gain come from, and is a correct workflow, once
generated, the one that is returned?

Comparisons of test-time inference methods commonly rest on two
assumptions: that a measured gain reflects the mechanism named by the
method's label, and that final accuracy adequately summarizes a method.
These issues remain insufficiently examined in
structured workflow planning. First, the
usual safeguard for attribution is a compute-matched baseline
\citep{wang2024reasoning}, but equal token expenditure is only a proxy.
Methods with the same budget can still differ in prompts, sampling
policies, candidate retention, post-processing, and serving behavior,
and a method's \emph{nominal} procedure can diverge from the
\emph{realized} procedure that actually runs. Second, final accuracy
conflates two failures: not \emph{discovering} a correct workflow and
failing to \emph{select} one that is already available. Verifier-based
inference separates generation from selection
\citep{cobbe2021training,lightman2024let}, and candidate coverage can
keep improving after selection saturates
\citep{brown2024large}; what remains unclear is where, between
generation and the final answer, correct candidates are lost.

In this paper, we examine both assumptions on DisasterBench
\citep{chen2026disasterbench}, which evaluates disaster-response
planning over 26 specialized tools with fixed interfaces. Its typed
workflow representation makes each stage of inference observable:
candidates can be compared exactly, structural validity can be checked
independently of task correctness, and tool- and dependency-level errors
remain visible. We compare Independent Sampling (IS), a nominal CoT-tree
configuration that realizes non-adaptive complete-workflow
generation, and multi-operator search (MO-Tree) on
GPT-4o-mini and Qwen3.5-9B under controlled inference
budgets. All candidate pools pass through a common pipeline that records
separately whether a correct workflow is \emph{discovered}, survives
\emph{structural validation}, and is \emph{selected}: a stage-wise
view we call the Discovery--Validation--Selection (DVS)
decomposition (Figure~\ref{fig:overview}).

To examine these questions, we introduce a
\emph{prompt-matched independent-sampling control}
that uses CoT-tree's expansion prompt without constructing
a search tree. We then trace candidate workflows through
the common evaluation pipeline to distinguish failures
in discovery, structural validation, and final selection.

Our analysis yields three main findings:
\begin{enumerate}
    \setlength{\itemsep}{0.2em}
    \setlength{\parsep}{0pt}
    \setlength{\parskip}{0pt}

    \item \textbf{Nominal tree search can obscure the realized
    generation mechanism.}
    Our CoT-tree configuration samples complete workflows
    non-adaptively. Independent sampling with the same
    expansion prompt reproduces its discovery coverage at
    comparable inference cost (154 vs.\ 153 of 183 tasks
    on GPT-4o-mini).

    \item \textbf{Correct workflows are lost at selection, not validation.}
    Using the DVS decomposition, we find that structural filtering
    never removes the last correct workflow, while majority voting
    leaves one unselected on 18--34 tasks per pool. In 57 of 60
    audited failures, the winning alternatives contain substantive
    tool-choice or information-flow errors.

    \item \textbf{Semantic verification does not reliably close the gap.}
    A tool-specification-aware verifier fixes 90 majority errors
    but overturns 175 correct decisions.
\end{enumerate}

We further distill the pipeline factors that can alter such
comparisons into an attribution checklist
(Table~\ref{tab:attribution-checklist}).
Code and reproducibility artifacts are publicly available at
\href{https://github.com/TamuChen18/Found-but-Not-Chosen}{GitHub}.

\begin{table*}[t]
    \centering
    \small
    \setlength{\tabcolsep}{3.5pt}
    \begin{tabular}{llrrrrrrr}
        \toprule
        Model / Run & Procedure &
        Oracle & Selected & Gap &
        Filter & Vote & Tie &
        Tokens/task \\
        \midrule

        GPT-4o-mini
        & IS
        & 139 & 121 & 18
        & 0 & 18 & 0
        & 238k \\

        & CoT-tree
        & 153 & 131 & 22
        & 0 & 22 & 0
        & 242k \\

        & MO-Tree
        & 146 & 123 & 23
        & 0 & 21 & 2
        & 239k \\

        & Prompt-matched IS$^\dagger$
        & 154 & 129 & 25
        & 0 & 24 & 1
        & 242k \\

        \midrule

        Qwen3.5-9B Control
        & Prompt-matched IS$^\dagger$
        & 160 & 127 & 33
        & 0 & 32 & 1
        & 253k \\

        \midrule

        Qwen3.5-9B Run 1
        & IS
        & 159 & 129 & 30
        & 0 & 28 & 2
        & 263k \\

        & CoT-tree
        & 162 & 132 & 30
        & 0 & 28 & 2
        & 253k \\

        & MO-Tree
        & 147 & 122 & 25
        & 0 & 19 & 6
        & 285k \\

        \midrule

        Qwen3.5-9B Run 2
        & IS
        & 162 & 128 & 34
        & 0 & 33 & 1
        & 263k \\

        & CoT-tree
        & 157 & 133 & 24
        & 0 & 22 & 2
        & 252k \\

        & MO-Tree
        & 147 & 122 & 25
        & 0 & 18 & 7
        & 284k \\

        \bottomrule
    \end{tabular}

    \caption{
        Results on 183 evaluation tasks. Prompt-matched IS matches
        CoT-tree's Oracle coverage; all pools lose correct workflows
        on 18--34 tasks at selection, with none lost to structural filtering.
        Gap = Oracle $-$ Selected; Filter, Vote, and Tie decompose
        the gap. Tokens/task is realized inference expenditure.
        $^\dagger$Prompt-matched IS rows are attribution
controls rather than primary generation procedures.
    }
    
    \label{tab:main-results}
\end{table*}

\section{Related Work}
\label{sec:related-work}

\paragraph{Test-Time Search, Verification, and Evaluation.}
Test-time inference spans repeated sampling and self-consistency
\citep{wang2022self}, verifier-based selection
\citep{cobbe2021training,lightman2024let}, structured search such as
Tree of Thoughts and RAP
\citep{yao2023tree,hao2023reasoning}, and MCTS over heterogeneous
reasoning actions~\citep{wu2024beyond}. Model-based evaluation provides
another selection mechanism, with LLM-as-a-judge methods using strong
language models to score or rank candidate outputs
\citep{zheng2023judging}. Recent work studies hybrid sampling
and search, inference-time budget allocation, and discriminator-guided
inference
\citep{chen2024tree,liu2026bag,suh2026think,fang2026inference}.
Budget-aware comparisons show that the apparent advantage of elaborate
reasoning strategies can vanish against strong compute-matched
baselines~\citep{wang2024reasoning,hariri2026testtime}; studies of agent
frameworks likewise show that implementation and evaluation choices can
materially affect measured performance
\citep{zhu2025oagents,yao2026harness}. Related work on multi-agent debate analyzes candidate
diversity and aggregation separately
\citep{zhu2026demystifying}. ToolPRMBench shows that weak process-reward models can harm tool-use
selection~\citep{li2026toolprmbench}, while generative self-refinement
offers an alternative to selecting directly from sampled candidates
\citep{wang2026learning}. We extend these observations to typed workflow
planning by separating discovery from selection and testing whether
apparent search gains survive scaffold-matched controls.

\paragraph{Disaster-Response AI and Workflow Planning.}
Prior work has studied disaster-specific capabilities including
remote sensing, flood mapping, damage assessment, and
language-model-based disaster analysis
\citep{gupta2019creating,bonafilia2020sen1floods11,
cui2026large,wang2026maestro}. Human--LLM collaboration has also been
used in disaster-domain benchmark construction and evaluation, including
DisastQA~\citep{chen2026disastqa}. DisasterBench
\citep{chen2026disasterbench} shifts this setting to typed multi-tool
workflow planning and diagnoses tool-, parameter-, and dependency-level
errors within individual structured plans. We build on that benchmark
to study multi-candidate test-time inference, asking where correctness
is lost when multiple candidate workflows are generated and one must
ultimately be selected. Extended related work and positioning
are provided in Appendix~\ref{app:related-work}.

\section{Task Setting and Evaluation Protocol}
\label{sec:protocol}

\subsection{Structured Disaster Planning}
\label{sec:task-setting}

We evaluate on DisasterBench~\citep{chen2026disasterbench}, which maps
natural-language disaster-response objectives to structured workflows
over 26 tools with fixed input/output interfaces. Each workflow
specifies tool invocations, task-specific inputs, and cross-step
dependencies. We use the first 50 of 233 expert-validated tasks
for development and the remaining 183 for evaluation. Two evaluation tasks informed GPT-4o-mini generation-count
calibration, while five informed a separate MO-Tree width
calibration. Excluding the union of these five tasks yields
similar conclusions
(Appendix~\ref{app:calibration-exclusion}).

We distinguish \emph{planning correctness} from \emph{structural
validity}. Planning correctness concerns appropriate tools,
parameters, and information-flow dependencies, whereas structural
validity concerns consistency with the workflow representation and
declared tool interfaces. A structurally valid workflow may therefore
still contain incorrect planning decisions. The workflow schema and
validation rules are detailed in Appendix~\ref{app:benchmark}.

\begin{figure}[t]
    \centering
    \includegraphics[width=\columnwidth]{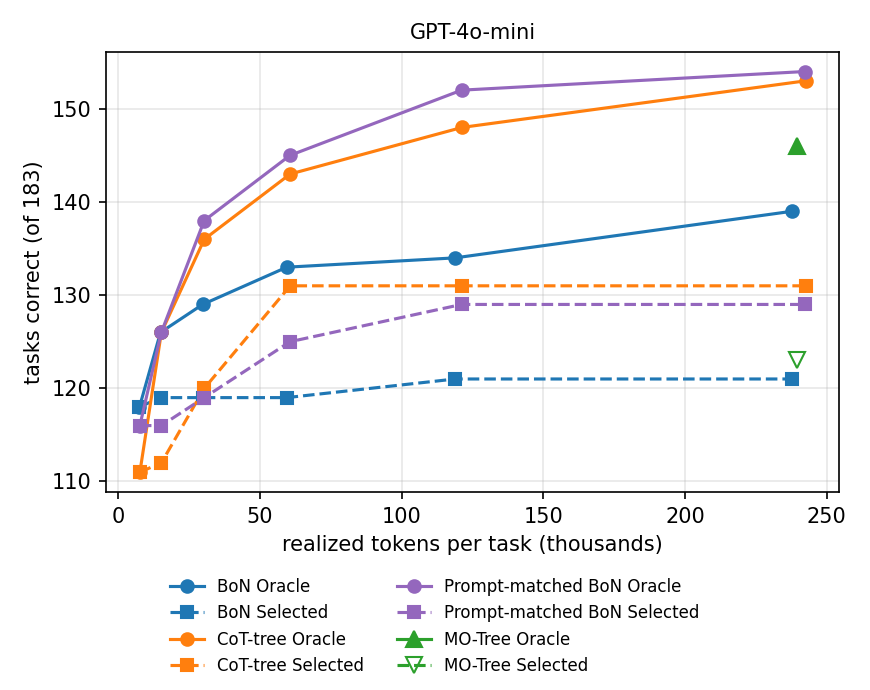}
    \caption{
    GPT-4o-mini generation-budget scaling.
    Prompt-matched IS tracks CoT-tree Oracle coverage from four generations onward.
    }
    \label{fig:budget-scaling}
\end{figure}

\subsection{Candidate Generation and Attribution Controls}
\label{sec:generation}

We compare three candidate-generation procedures using a shared task
description, tool manifest, few-shot demonstrations, and output format.
\emph{IS} samples complete workflows independently using the baseline
Chain-of-Thought (CoT) prompt. \emph{CoT-tree} is instantiated from an
existing MCTS implementation for multi-operator reasoning
\citep{qi2025mutual,wu2024beyond} as its CoT-only variant: the other
four reasoning operators are disabled, and each CoT sample is issued
as a separate request and retained as an independent leaf. At the
complete-workflow level, this study-specific configuration is
non-adaptive because later requests do not condition on previously
generated workflows.

To isolate whether CoT-tree's observed discovery advantage requires
its search structure, we introduce a \emph{prompt-matched
independent-sampling control} (prompt-matched IS). It uses the same
expansion prompt, decoding settings, and generation count as CoT-tree,
but samples complete workflows independently without constructing
a search tree. We additionally test CoT-tree's retention convention
on the control in Section~\ref{sec:main-results}.
IS, CoT-tree, and prompt-matched IS use 32 generations per task
for GPT-4o-mini and 34 for Qwen3.5-9B. These counts were
calibrated against average MO-Tree request counts in preliminary
runs, before the final generation configurations were fixed.
Realized request counts and inference costs are reported in
Appendix~\ref{app:search-budget}.

\emph{MO-Tree} combines rephrasing, direct answering, Chain-of-Thought,
divide-and-conquer, and self-refinement, and serves as the adaptive
multi-operator search comparator. Unlike CoT-tree, later expansions
may condition on intermediate search states. Its realized branching
depends on the serving backend (Appendix~\ref{app:backend}).

We measure inference budgets using realized token expenditure within
each model and serving backend. Full generation configurations,
calibration, decoding settings, retention rules, backend behavior, and
resource accounting are detailed in
Appendix~\ref{app:search-budget}.

\subsection{Common Post-Generation Evaluation}
\label{sec:matched-eval}

Figure~\ref{fig:overview} summarizes the generation and evaluation
pipeline. All candidate pools undergo the same offline stages:
\emph{extraction}, \emph{normalization}, \emph{structural validation},
and \emph{selection}. Normalization permits only representation-level
transformations uniquely determined by the generated workflow and fixed
tool interfaces; transformations that discard or overwrite generated
planning decisions are disabled. Structural validation checks internal
consistency with the workflow representation and tool interfaces,
without evaluating task-specific planning correctness. These are
representation-level checks; input values and types are not validated,
so a candidate can pass all checks while binding a downstream input to
an unsuitable upstream product. Rules and ablations appear in
Appendices~\ref{app:transformation-audit} and~\ref{app:benchmark}.

For primary selection, normalized workflows are grouped by exact
workflow identity. We use plurality voting (often referred to as
majority voting in LLM self-consistency): the workflow group with
the highest retained vote mass is selected, with deterministic
tie-breaking. An absolute majority is not required. Vote multiplicity follows
each procedure's native retention semantics: IS retains repeated
generations, CoT-tree retains deduplicated complete outputs, and MO-Tree
retains aggregated solution nodes. Thus, the selector is shared while
candidate multiplicity remains procedure-dependent. If all candidates are filtered out, selection falls back
to the normalized unfiltered pool; this fallback is never
triggered in the 2,013 evaluated pool--task cases.
Alternative conventions and selectors are examined in
Appendix~\ref{app:selector-sensitivity}.

\subsection{Evaluation Measures}
\label{sec:metrics}

We measure planning correctness using strict Exact Match (EM) against
expert-validated reference workflows. \emph{Oracle EM} measures whether
at least one normalized candidate matches the reference before
structural filtering, whereas \emph{Selected EM} measures whether the
final selected workflow matches the reference. The
\emph{Oracle--Selected gap} counts tasks for which a correct workflow is
discovered but not returned.

We refer to this stage-wise evaluation as the
\emph{Discovery--Validation--Selection (DVS) decomposition}.
It distinguishes failures to discover a correct workflow
from losses during structural validation and final selection.
For each task $t$, let $O_t$ and $S_t$ indicate Oracle and
Selected correctness. Let $F_t$, $V_t$, and $E_t$ indicate
filter, vote, and tie losses, respectively. All five
variables are binary indicators, with $S_t \leq O_t$
because the selected workflow is drawn from the
normalized candidate pool. Under our evaluation protocol,
these mutually exclusive loss indicators satisfy
\begin{equation}
O_t-S_t=F_t+V_t+E_t.
\label{eq:dvs}
\end{equation}
A filter loss occurs when structural validation removes the
last correct candidate; a vote loss occurs when an incorrect
workflow receives strictly more support than every surviving
correct workflow; and a tie loss occurs when a correct workflow
reaches the highest vote count but loses the deterministic
tie-break. Summing over tasks yields the Gap, Filter, Vote,
and Tie counts in Table~\ref{tab:main-results}.

We also examine candidate invalidity and pool-level
characteristics. Aggregated counts describe pool--task cases, not
independent observations; paired tests are confined to matched
within-pool comparisons. Paired comparisons use exact two-sided
McNemar tests and paired bootstrap confidence intervals, with
Qwen3.5-9B runs analyzed separately; full results appear in
Appendix~\ref{app:stat-tests}.

\section{Experiments and Main Results}
\label{sec:experiments}

\subsection{Does CoT-tree's Discovery Gain Survive Prompt Matching?}
\label{sec:main-results}

\textbf{Prompt-matched independent sampling reproduces
CoT-tree's discovery advantage.}
On GPT-4o-mini, CoT-tree discovers correct workflows on 153
of 183 tasks, compared with 139 for IS
(Table~\ref{tab:main-results}). Yet prompt-matched IS reaches
154 at comparable inference cost, differing from CoT-tree by
only one Oracle-correct task ($p=1.000$) while exceeding IS
by 15 tasks ($p=.0015$). The two candidate pools also show
similar diversity and winning vote concentration
(Table~\ref{tab:app-pool-stats}). Applying CoT-tree's
retention convention to the control leaves both Oracle and
Selected counts unchanged. The observed CoT-tree advantage
over IS therefore provides no evidence for an additional
tree-structure effect.

This attribution result is robust to normalization. Before common
normalization, CoT-tree and prompt-matched IS achieve Oracle counts
of 145 and 148, respectively, compared with 139 for IS; afterward,
they reach 153 and 154, while IS remains at 139. Thus, normalization
changes the observed advantage over baseline IS but not the finding
that scaffold-matched independent sampling matches or exceeds
CoT-tree (Appendix~\ref{app:transformation-audit}).

\textbf{Prompt-matched sampling also matches CoT-tree-level
discovery on Qwen3.5-9B.}
Prompt-matched IS discovers correct workflows on 160 tasks,
compared with 162 and 157 in the two CoT-tree runs
(Table~\ref{tab:main-results}). Its Selected count is 127,
versus 132 and 133 for CoT-tree; these paired differences
are not statistically significant. Applying CoT-tree's
retention convention increases the control's Selected count
to 130, while candidate diversity remains comparable
(Appendix~\ref{app:search-budget}).
Thus, prompt-matched IS falls within the observed
CoT-tree run range in Oracle coverage, supporting the
same attribution conclusion across both models.

\textbf{No generation procedure consistently outperforms
independent sampling.}
On GPT-4o-mini, MO-Tree does not differ significantly
from IS. On Qwen3.5-9B, CoT-tree remains within five Oracle-correct
tasks of IS in both runs, whereas MO-Tree has significantly lower
Oracle coverage ($p\leq .008$). Restoring MO-Tree's original search
width increases inference cost by 54\% but recovers only three
Oracle-correct tasks (Appendix~\ref{app:qwen-width}). A separate 40-task pilot with active reward-guided UCT
selection likewise finds no statistically significant
accuracy advantage over random tree selection or
independent sampling at approximately matched inference
cost (Appendix~\ref{app:active-uct}). Because MO-Tree
realizes different local branching behavior across the two serving
backends, we interpret cross-model MO-Tree comparisons descriptively.
Detailed paired statistics are provided in
Appendix~\ref{app:stat-tests}. Across all eleven candidate pools,
Oracle EM exceeds Selected EM by 18--34 tasks, showing that discovering
a correct workflow does not ensure its selection;
Section~\ref{sec:analysis} examines these post-discovery losses.

\subsection{How Do Discovery and Selection Scale with Budget?}
\label{sec:budget-scaling}

To examine how discovery and selection change with generation budget,
we reconstruct chronological prefixes of IS, prompt-matched IS, and
CoT-tree outputs. Because complete-workflow
generations are non-adaptive for these procedures, prefixes of
$N=1,2,4,8,16$ provide valid post-hoc sampling conditions.
Adaptive MO-Tree is evaluated only at its complete-run operating
points.

\textbf{Candidate discovery continues to improve after selected
accuracy plateaus.}
Figure~\ref{fig:budget-scaling} shows this pattern on GPT-4o-mini. CoT-tree reaches its full-budget Selected count
of 131 correct tasks by eight generations, while Oracle coverage
continues to increase to 153. Prompt-matched IS closely tracks this
discovery curve: at eight generations it reaches 145 Oracle-correct
tasks versus 143 for CoT-tree, compared with 133 for IS. Thus, the
attribution result is not confined to the full-budget endpoint:
independent sampling under the CoT-tree generation scaffold reproduces
the same discovery advantage over IS from four generations onward.

\textbf{Selection saturation is not unique to GPT-4o-mini.}
On Qwen3.5-9B Run 1, IS Oracle coverage rises from 107 correct tasks after one
generation to 159 at the full budget, whereas Selected accuracy remains
between 124 and 129 correct tasks from four generations onward.
Overall, additional sampling can continue improving candidate discovery
after gains in final selection accuracy have largely plateaued.
Full scaling curves for both models and a complementary random
candidate-subsampling analysis are provided in
Appendices~\ref{app:budget-prefix} and~\ref{app:random-subsampling}.

\begin{table*}[t]
    \centering
    \small
    \setlength{\tabcolsep}{4pt}
    \begin{tabular*}{\textwidth}{
        @{\extracolsep{\fill}}
        p{0.18\textwidth}
        p{0.49\textwidth}
        p{0.27\textwidth}
        @{}}
        \toprule
        \textbf{Factor} &
        \textbf{Evidence in this study} &
        \textbf{Alignment principle} \\
        \midrule

        Generation scaffold &
        Prompt-matched independent sampling reaches CoT-tree-level
        Oracle coverage on GPT-4o-mini (154 versus 153) and Qwen3.5-9B
        (160 versus 162 and 157). &
        Match scaffolds or include a scaffold-matched sampling control.
        \\

        Sampling controls &
        In a matched IS rerun, removing the fixed request seed increases
        distinct raw responses from 2.03 to 3.96 per task and distinct valid
        workflows from 1.60 to 2.32; Oracle changes from 133 to 139
        ($p=.11$). &
        Apply decoding and seed policies symmetrically and distinguish matched
        controls from broader sensitivity checks.
        \\

        Post-generation processing &
        On earlier fixed-seed pools, common normalization reduces
        CoT-tree invalidity from 27.4\% to 2.9\% and increases
        unfiltered-majority correctness from 104 to 118.
        On primary GPT-4o-mini pools, CoT-tree Oracle coverage
        increases from 145 to 153, while IS remains at 139. &
        Use a common audited normalization policy.
        \\

        Backend realization &
        GPT-4o-mini reduces nominal multi-sample MO-Tree requests to
        single-sample updates, whereas Qwen realizes local aggregation. &
        Report realized rather than nominal branching.
        \\

        Resource accounting &
        A pooled candidate set estimated at 232k tokens/task from its
        retained subset actually costs 414k once both source runs are
        counted, approximately $1.74\times$ the IS budget. &
        Count all inference contributing to the evaluated pool.
        \\

        Selection interface &
        Generic and targeted LLM selectors do not reliably improve on
        majority selection. The targeted Top-3 verifier fixes 90 known
        errors but breaks 175 majority-correct decisions under
        unrestricted deployment; vote-margin gating removes most of
        this harm but yields no reliable gain. &
        Hold the primary selector fixed, preserve candidate-support
        information, and report selector sensitivity separately.
        \\

        \bottomrule
    \end{tabular*}
    \caption{
    Compact attribution checklist for structured test-time inference.
    A detailed version with detection criteria and implementation
    guidance is provided in Appendix~\ref{app:full-checklist}.
    }
    \label{tab:attribution-checklist}
\end{table*}

\section{Analysis}
\label{sec:analysis}

\subsection{Where Are Correct Workflows Lost?}
\label{sec:selection-gap}

\textbf{Validation preserves discovered correct workflows;
selection frequently loses them.}
Across all eleven candidate pools, structural filtering never
removes the last correct workflow from an Oracle-positive task.
Of the 289 discovered-but-unselected pool--task cases,
265 (91.7\%) are strict vote losses and the remaining 24
are tie losses. This zero filter-loss count is empirical
rather than guaranteed by the validation procedure:
none of the exact-match candidates among 45,246 normalized
candidates fails structural validation. Filtering changes
the selected workflow in 16 pool--task cases, correcting
six selections without introducing any new errors.
Thus, filtering can affect selection without removing
correct candidates. The 18--34-task Oracle--Selected gaps
in every pool exceed the observed differences in Selected EM
across generation procedures, which are at most 11 tasks
within each model and run.

\textbf{Correct workflows often remain highly ranked.}
Among the 209 strict vote losses in the nine primary pools,
the best correct workflow ranks second in 57.9\% of cases.
We define \emph{Top-$k$ availability} as the number of tasks
with an exact-match workflow among the $k$ highest-ranked
workflow groups under the primary selector's ordering.
Top-1 availability therefore equals Selected EM.

Across the nine primary pools, a correct workflow appears
within the Top-2 groups for 137 of 231 Oracle--Selected
losses (59.3\%) and within the Top-3 groups for 181
(78.4\%). These are candidate-availability rates, not
achieved correction rates. Availability varies by procedure:
for GPT-4o-mini CoT-tree, ten Oracle-positive tasks have
no correct workflow among the top three groups.

These results do not reflect trivially small candidate pools:
strict vote-loss tasks contain a median of five distinct
valid workflow groups, compared with two across all tasks.
Many selection failures therefore require discriminating
among several plausible workflow alternatives
(Appendix~\ref{app:topk-selection}).

\subsection{What Errors Win the Vote?}
\label{sec:error-audit}

\textbf{Most winning vote-loss errors reflect substantive
planning mistakes rather than Exact-Match artifacts.}
To distinguish genuine planning errors from representation-level
differences, we audit 60 non-tie vote-loss cases, sampling ten
from each model--procedure pool for GPT-4o-mini and
Qwen3.5-9B Run 1. Initial labels are produced by an LLM
annotator under a fixed rubric, and three human reviewers
examine all 60 cases with access to the LLM-generated labels.
All three confirm the error-category and semantic-equivalence
labels without revision. This constitutes LLM-assisted human
review rather than independent blind annotation
(Appendix~\ref{app:error-audit}).

Of the 60 audited cases, 57 (95.0\%; Wilson 95\% CI:
86.3--98.3\%) contain substantive planning errors: 35 involve
incorrect tool selection and 22 involve incorrect dependency
construction. The remaining three are semantics-preserving
permutations of independent workflow steps. An automatic
permutation-equivalence check identifies all three without flagging
any of the 57 substantive errors. Thus, the audited selection failures
predominantly reflect incorrect planning decisions rather than
strict-EM artifacts. For example, on Task~123, the winning workflow substitutes
\texttt{Building\_damage\_assessment} for
\texttt{Change\_Mapping\_and\_Detection}, receiving 22 votes
against two for the correct workflow
(Appendix~\ref{app:error-audit}).

\textbf{High vote support does not guarantee correct
planning decisions.}
Across all 265 strict vote-loss cases, only 27.5\% of
incorrect winners preserve the reference tool sequence,
and none achieve full parameter correctness.
High-margin incorrect winners preserve the reference
tool sequence more often than low-margin winners
(36.4\% versus 18.8\%), suggesting that strong vote
support can favor workflows with correct tool choices
but incorrect downstream decisions.
Of the 73 winners with the correct tool sequence,
68 contain incorrect cross-step input bindings,
while the remaining five differ only in literal
input values (Appendix~\ref{app:vote-loss-partial}).

\paragraph{A concrete selection failure.}
Task~22 illustrates how majority voting can favor a
structurally valid but incorrect dependency structure.
The correct workflow contains independent downstream
branches, whereas the winning alternative introduces
a sequential dependency between them. Under GPT-4o-mini
IS, this incorrect workflow receives 25 of 32 votes,
despite a correct workflow being available in the
candidate pool. The same incorrect workflow receives
only one vote under prompt-matched IS and two under
CoT-tree (Appendix~\ref{app:task22-case}).

\subsection{Do Winning Errors Recur?}
\label{sec:modal-errors}

\textbf{Incorrect majority winners can recur across independent
runs and different generation procedures.}
The degree of recurrence varies with differences in candidate
distributions across procedures. On Qwen3.5-9B,
where CoT-tree and IS reach similar Oracle coverage, they select the
same incorrect workflow on 14 of 15 shared vote-loss tasks (93\%),
comparable to IS's own cross-run recurrence (24 of 25, 96\%). On
GPT-4o-mini, where CoT-tree and IS produce more distinct candidate
distributions, IS repeats CoT-tree's incorrect winner on only 5 of
CoT-tree's 22 vote-loss tasks (23\%), and the mean winning
wrong-workflow share falls from 68\% under IS to 41\% under CoT-tree.
GPT-4o-mini IS errors are concentrated: in only one of 18 vote-loss
cases is the best correct workflow within two votes of the incorrect
winner. Further recurrence statistics are provided in Appendix~\ref{app:modal-errors}.

\paragraph{Why does majority voting fail?}
Taken together, these results suggest that repeated sampling
can reinforce recurring planning errors rather than eliminate
them through voting. Incorrect winners often receive strong
support and can recur across runs, while many contain subtle
tool-choice or cross-step dependency errors despite being
structurally valid. Because majority voting rewards frequency
rather than semantic correctness, a repeatedly generated
incorrect workflow can defeat an available correct alternative.

\subsection{Can Semantic Verification Close the Selection Gap?}
\label{sec:selector-sensitivity}

\textbf{Majority voting remains a competitive selection baseline.}
Removing structural filtering or changing the vote-counting
convention changes Selected accuracy by at most five tasks
per primary pool, and an oracle tie-break recovers at
most seven, whereas selecting the first valid candidate loses up to 19,
indicating that candidate frequency carries useful signal. A generic
GPT-4o judge changes Selected accuracy by $+6$ to $-22$ tasks and
reverses the IS--MO-Tree ordering on both Qwen runs; because it sees
only a bounded, deduplicated subset of candidates, we interpret this
as sensitivity to the evaluated judge interface rather than evidence
against LLM-based verification in general
(Appendix~\ref{app:selector-sensitivity}).

\textbf{Targeted semantic verification fixes some majority
errors but overturns more correct decisions.}
A targeted, tool-specification-aware Top-3 verifier provides
a more focused diagnostic. Its calls are excluded from
candidate-generation budgets but accounted for separately. On the 181 pool--task cases in the nine primary
pools where majority selection is incorrect but a correct workflow
ranks in the top three, it selects correctly in 90 (49.7\%), versus 66 expected correct cases under uniform random choice. Across all
contested cases with at least two valid workflow groups, it fixes
90 majority errors but overturns 175 correct majority decisions;
eight of nine pools decline, four significantly.

\textbf{Restricting verifier overrides by vote margin does not
produce reliable gains.}
False overrides occur in every vote-margin bin, but because
most majority-correct cases have high margins, the harm
concentrates there;
false overrides also add steps more often than fixes do (52.6\% versus
27.8\%). Permitting an override only when the normalized margin between
the two highest-ranked groups is at most a threshold $t$ removes much
of this harm but yields no reliable gain: the best post-hoc threshold
($t=.15$) adds four pool--task cases ($p=.69$), and leave-one-run-out
threshold selection yields a net change of $-1$. Neither unrestricted
nor margin-gated verification reliably improves upon majority
selection (Appendix~\ref{app:targeted-verifier}). \paragraph{Implications for selector design.}
These results suggest three directions for future selectors:
checking cross-step semantic compatibility, accounting for
candidate support when overriding majority decisions, and
avoiding unsupported preferences for longer workflows.

\subsection{Attribution Checklist for Structured Test-Time Inference}
\label{sec:pipeline-sensitivity}

Table~\ref{tab:attribution-checklist} consolidates the pipeline
factors that can affect comparisons between inference
procedures, together with the corresponding controls
or reporting principles.

Although our experiments focus on disaster workflow planning, the
identified confounds are not inherently domain-specific. Test-time
inference procedures may differ in generation scaffolds, sampling
policies, post-processing, realized branching, resource accounting,
and selection interfaces. We therefore recommend reporting
scaffold-matched controls where applicable, realized rather than
nominal compute and branching, and Oracle and Selected accuracy
separately. These measures help distinguish candidate-discovery
differences from selection effects.

\section{Conclusion}

We investigate attribution and selection in structured test-time
inference for disaster-response workflow planning. The apparent
advantage of a nominal tree-search configuration is reproduced by
prompt-matched independent sampling, showing that method labels alone
are insufficient to explain observed gains and that comparisons should
be grounded in the realized inference pipeline. We further identify a
persistent gap between discovery and selection: correct workflows can
remain available after structural validation yet lose to plausible
incorrect alternatives during final selection. A targeted semantic
verifier recovers some of these failures but does not reliably improve
upon majority selection, and unrestricted use overturns more correct
decisions than it fixes. These findings motivate scaffold-matched attribution controls
and the DVS decomposition for distinguishing candidate
discovery, structural validation, and final selection.
For disaster-response planning, this distinction is
particularly important: structurally valid workflows
can still contain incorrect tool choices or information-flow
dependencies, potentially undermining downstream analyses.

\section*{Limitations}

Our study uses one benchmark, DisasterBench, and two model families.
Evaluation across additional models and structured-planning benchmarks
would clarify how broadly the findings transfer. DisasterBench also
evaluates offline workflow construction, so runtime execution effects
are outside our scope.

We do not claim that adaptive tree search is ineffective. The evaluated
CoT-tree configuration is non-adaptive at the complete-workflow level,
and MO-Tree realizes different branching behavior across serving
backends. Our claims therefore concern attribution in these realized
pipelines rather than the general efficacy of tree search.

Our semantic audit covers 60 sampled vote-loss cases, with LLM-assisted
labels reviewed by three human reviewers, and our equivalence check
targets permutations of independent steps. Our selection analysis covers
majority-based rules and LLM-based judges; learned, execution-aware,
or confidence-calibrated selectors may recover more of the available
Oracle coverage.

\section*{Ethics Statement}
\label{sec:ethics}

This study investigates test-time inference and candidate
selection in structured disaster-response workflow planning.
All experiments are conducted offline using the DisasterBench
benchmark and predefined tool interfaces. The study does not
involve real-world emergency operations, human participants,
personally identifiable information (PII), or private
operational records.

Although the evaluated workflows are motivated by
disaster-response applications, the results should not be
interpreted as evidence that the studied planning methods
are suitable for autonomous deployment in safety-critical
environments. In particular, structurally valid workflows
may still contain incorrect tool selections, parameter
bindings, or dependencies, and improvements in candidate
generation or oracle performance do not necessarily
translate into reliable final decisions.

Our findings highlight the importance of transparent
evaluation protocols, consistent computational accounting,
and reliable candidate selection when assessing
search-based LLM planning methods. We encourage future
research toward execution-grounded evaluation, rigorous
safety validation, and human oversight before applying
such systems to real-world disaster-response operations.

\section*{Acknowledgments}

We thank Ankit Basu for his contributions to an early
prototype of the MCTS pipeline.

This work used the ACES at Texas A\&M University,
DeltaAI, and Delta GPU resources at the National Center
for Supercomputing Applications through allocations
CIV250019 and CIV250021 from the Advanced
Cyberinfrastructure Coordination Ecosystem:
Services \& Support (ACCESS) program, which is
supported by U.S. National Science Foundation grants
\#2138259, \#2138286, \#2138307, \#2137603,
and \#2138296.

We also acknowledge the use of high-performance
computing resources provided by the Texas A\&M
University High Performance Research Computing
(HPRC) facility.
\clearpage
\bibliography{custom}

\clearpage
\appendix

\section{Real-World Disaster-Response Grounding}
\label{app:real-world-grounding}

Although DisasterBench tasks are synthetic, their multi-stage
structure reflects information-processing requirements found in
real disaster response: selecting analytical capabilities, binding
task-specific inputs, and passing intermediate products between
stages. We illustrate these parallels using two documented events.
The examples are operational analogies, not reconstructions of
the events using DisasterBench tools.

\paragraph{Hurricane Harvey (2017).}
During Hurricane Harvey, NASA/JPL produced flood proxy maps of
southeastern Texas using pre- and post-event SAR observations
from ALOS-2 and Sentinel-1~\citep{nasa2017harvey}.
Subsequent research demonstrated Sentinel-1 coherence-based
detection of floodwater in urban Houston~\citep{chini2019sentinel}.
These activities illustrate an information-processing chain from
selecting imagery for the affected region and dates, through
sensor-appropriate flood mapping, to producing geospatial
information for downstream response support. In DisasterBench,
analogous planning requirements involve choosing appropriate
tools and preserving references to upstream analytical outputs
across workflow steps.

\paragraph{Maui Wildfires (2023).}
Following the Maui wildfires, pre- and post-event ALOS-2 SAR
observations were used to generate damage proxy maps over
Lahaina. These products supported disaster assessment and
response activities~\citep{rolla2025satellite}. This illustrates
dependencies between imagery selection, change analysis, and
damage assessment, resembling the analytical roles of
\texttt{Change\_Mapping\_and\_Detection} and
\texttt{Building\_damage\_assessment} in DisasterBench.
An incorrect upstream image or parameter can compromise a
downstream analysis even when the resulting workflow satisfies
interface-level structural constraints. Our evaluation studies
such planning correctness at the schema level, without executing
the tools or validating real-world response outcomes.

\section{Benchmark and Workflow Representation}
\label{app:benchmark}
\label{app:validity}

This appendix summarizes the workflow representation and
representation-level validity checks inherited from
DisasterBench~\citep{chen2026disasterbench} and used in our
evaluation pipeline.

\subsection{Tool Manifest and Workflow Schema}

DisasterBench defines 26 disaster-response tools with fixed
input/output interfaces. In the released data and evaluator, the tool
invoked at each step is stored in a field named \texttt{agent}; we use
``tool'' throughout the paper for readability. Each tool specification
contains a natural-language description (\texttt{desc}) and input
(\texttt{input}) and output (\texttt{output}) maps that define the
available interface fields.

A workflow is an ordered sequence of tool invocations. Each step
selects a tool and binds its inputs either to literal task values or to
outputs produced by earlier steps. The normalized representation used
by the evaluator contains six fields per step:

\begingroup
\small
\begin{verbatim}
[
  {
    "agent": "<tool_name>",
    "step": 0,
    "dependence": [-1],
    "dependence_content": null,
    "inputs": {
      "<input_key>": "<literal_value>"
    },
    "outputs": ["<output_key>"]
  },
  {
    "agent": "<tool_name>",
    "step": 1,
    "dependence": [0],
    "dependence_content": {"0": ["<output_key>"]},
    "inputs": {
      "<input_key>":
        "<GENERATED>-0-<output_key>"
    },
    "outputs": ["<output_key>"]
  }
]
\end{verbatim}
\endgroup

Step identifiers are explicit through \texttt{step} and coincide with
list position in the reference workflows. Steps without predecessors
use \texttt{dependence} $=[-1]$. Exact Match performs an ordered deep
comparison over the normalized step objects, including all six fields.
Because the comparison is ordered, permutations of independent steps
are counted as mismatches; Section~\ref{sec:error-audit} quantifies how
often this occurs among sampled selection failures.

Cross-step information flow is encoded through symbolic references to
outputs produced by earlier steps. A generated-output reference
occupies the complete input value and uses a zero-based source-step
index followed by the output key, for example:

\begingroup
\small
\begin{verbatim}
<GENERATED>-0-<high_res_image_T_prime_path>
\end{verbatim}
\endgroup

More generally,

\begingroup
\small
\begin{verbatim}
<GENERATED>-<source_step>-<output_key>
\end{verbatim}
\endgroup

The tool manifest is used as an interface specification rather than as
an oracle for correcting generated planning decisions. In particular,
E4, the manifest-assisted transformation that replaces an erroneous
generated output-key choice, is disabled in the primary protocol
because it can overwrite a model-generated planning decision
(Appendix~\ref{app:transformation-audit}).

\subsection{Dependency Representation and Structural Validity}

Workflow dependencies are represented through
\texttt{dependence} and \texttt{dependence\_content}. During
normalization, these fields are reconstructed from generated-output
references in the step inputs. The dependency structure used by the
evaluator therefore reflects the information flow expressed by the
generated workflow rather than the model's original explicit
dependency fields.

If Step~$j$ consumes an output generated by Step~$i$, the normalized
workflow contains the dependency $i \rightarrow j$.

With E2b disabled---the transformation that would otherwise discard
additional predecessors from a multi-reference dependency
(Appendix~\ref{app:transformation-audit})---all recovered predecessor
references are retained in order of first occurrence. The validator
accepts only well-formed generated-output references; malformed
references are rejected under check~3 below. No reference workflow
contains an input with multiple embedded generated references, and no
reference step contains more than one predecessor.

Using the notation of the Task~22 example, the representation
distinguishes two downstream tools that independently consume the same
upstream output,

\[
X_{\mathrm{img}} \rightarrow T_{\mathrm{obj}},
\qquad
X_{\mathrm{img}} \rightarrow T_{\mathrm{land}},
\]

from a sequential dependency,

\[
X_{\mathrm{img}} \rightarrow T_{\mathrm{obj}}
\rightarrow T_{\mathrm{land}}.
\]

Both structures can be representation-level valid while encoding
different planning decisions.

Structural validation checks whether a normalized workflow is
internally consistent with the workflow representation and declared
tool interfaces. The primary validator applies the following checks:

\begin{enumerate}
    \setlength{\itemsep}{0.15em}

    \item The workflow must parse as a non-empty JSON list of steps,
    and every invoked tool must exist in the manifest.

    \item Every input key declared by the invoked tool must be present.
    This is a presence check only: input values and types are not
    validated, and additional input keys are not penalized.

    \item Every generated-output reference must be well formed and
    refer to a strictly earlier step, rejecting malformed,
    self-referential, and forward references.

    \item Every referenced output key must be declared either in the
    source tool's manifest or in the source step's explicit
    \texttt{outputs} field.
\end{enumerate}

Normalization rebuilds \texttt{dependence} from the generated-output
references before validation, so the validator's
dependency--reference consistency checks are effectively redundant on
the primary path. Across all 45,246 candidates in the eleven pools,
they fire on a single candidate, which is also rejected for a
malformed reference, and never determine a rejection on their own.
\texttt{dependence\_content} is not checked by the validator.

A candidate enters the structurally filtered pool only when it incurs
zero hard violations. Across the eleven evaluated pools, 42,127 of
45,246 candidates pass structural validation. Acyclicity follows from
the requirement that generated references point only to earlier steps,
so no separate cycle check is needed. The validator performs no
additional reachability check.

The output-key rule above is purely a validation operation: it does
not replace an erroneous key with a manifest-compatible one. Likewise,
the validator permits multiple predecessors when they are represented
consistently. These behaviors are distinct from the disabled E4 and
E2b transformations, which respectively overwrite an output-key choice
or discard predecessor information
(Appendix~\ref{app:transformation-audit}).

These checks establish representation-level validity rather than
task-level planning correctness. In particular, a candidate may pass
validation even when a downstream input is bound to an unsuitable
upstream product, provided that the reference is well formed and
interface-compatible at the representation level. This distinction
underlies the stage-wise analysis in Section~\ref{sec:selection-gap};
Appendix~\ref{app:task22-case} gives a concrete example in which both
dependency structures are representation-level valid but only one is
planning-correct.

\section{Search Configuration and Resource Accounting}
\label{app:search-budget}

This appendix documents the realized generation configurations,
backend behavior, and resource-accounting rules used in the
experiments. All reported inference costs include model inference
required to construct the evaluated candidate pool, including
generations that are later discarded, deduplicated, or aggregated.

\paragraph{Implementation provenance.}
Our MCTS implementation adapts search and utility components
from the open-source rStar framework~\citep{qi2025mutual},
with modifications for DisasterBench's typed workflow planning
setting. The released code retains the applicable upstream
copyright and license notices.

\subsection{Generation Configurations}
\label{app:generation-configs}

Table~\ref{tab:app-search-configs} summarizes the generation
configurations used in the experiments. For GPT-4o-mini, IS,
prompt-matched IS, CoT-tree, and MO-Tree all omit the request seed.
The first three use 32 generation requests per task and consume nearly
identical inference budgets. For Qwen3.5-9B, IS and CoT-tree use
34 generation calls per task, while MO-Tree uses a budget-calibrated
narrower configuration.

\paragraph{Generation-count and width calibration.}
The fixed sampling counts originate from preliminary MO-Tree
request-count calibration. For GPT-4o-mini, two evaluation
tasks yielded 35 and 28 requests, averaging 31.5 and giving
a rounded sampling count of 32. For Qwen3.5-9B, five evaluation
tasks yielded 31, 36, 34, 43, and 26 requests, averaging 34.0.
Both calibrations used earlier two-rollout MO-Tree configurations;
the Qwen calibration also used a different serving backend
from the final local vLLM evaluation.

Separately, five GPT-4o-mini evaluation tasks were used to
calibrate the narrower MO-Tree configuration, selecting
a two-rollout configuration with an average of 22.8 requests
per task. These calibration tasks are excluded in the
sensitivity analysis in Appendix~\ref{app:calibration-exclusion}.
Final compute comparisons use realized token expenditure
rather than assuming equal request counts imply equal cost.
\begin{table*}[t]
    \centering
    \scriptsize
    \setlength{\tabcolsep}{3pt}
    \begin{tabular}{llrrrrl}
        \toprule
        Model & Procedure &
        Calls/task &
        Retained/task &
        Tokens/task &
        Max output &
        Configuration \\
        \midrule

        GPT-4o-mini
        & IS
        & 32.0
        & 32.0
        & 238k
        & 2048
        & baseline prompt, no seed \\

        GPT-4o-mini
        & Prompt-matched IS
        & 32.0
        & 32.0
        & 242k
        & 2048
        & CoT-tree prompt, no seed \\

        GPT-4o-mini
        & CoT-tree
        & 32.0
        & 31.5
        & 242k
        & 2048
        & non-adaptive CoT expansion, no seed \\

        GPT-4o-mini
        & MO-Tree
        & 30.5
        & 12.7
        & 239k
        & 2048
        & five operators, no seed \\

        \midrule

        Qwen3.5-9B
        & IS
        & 34.0
        & 34.0
        & 263k
        & 2048
        & $N=34$ \\

        Qwen3.5-9B
        & Prompt-matched IS
        & 34.0
        & 34.0
        & 253k
        & 2048
        & CoT-tree prompt \\

        Qwen3.5-9B
        & CoT-tree
        & 34.0
        & 11.2
        & 253k
        & 2048
        & non-adaptive CoT expansion \\

        Qwen3.5-9B
        & MO-Tree
        & 33.6
        & 7.3
        & 285k
        & 2048
        & budget-calibrated \\

        \bottomrule
    \end{tabular}

    \caption{
    Generation configurations used in the experiments.
    Calls, retained candidates, and token usage are averaged per task.
    Prompt-matched IS rows are attribution controls rather than
    additional primary procedures.
    A request with $n>1$ counts as one backend request even when
    multiple completions are returned.
    Qwen3.5-9B token statistics are shown for Run~1; Run~2 uses the
    same generation configuration.
    }
    \label{tab:app-search-configs}
\end{table*}

\paragraph{Independent Sampling (IS).}

The IS procedure repeatedly samples complete workflows from the original task using
the baseline CoT prompt. Each generation is independent of previously
generated candidates.

\paragraph{CoT-tree.}
CoT-tree is instantiated from an MCTS implementation adapted
for DisasterBench's typed workflow planning setting, using
search and utility components derived from the open-source
rStar framework~\citep{qi2025mutual}. We evaluate a CoT-only
configuration that disables the other four reasoning operators
and retains each CoT sample as an independent leaf. In the realized configurations evaluated
here, complete-workflow generations are therefore non-adaptive rather
than conditioned on earlier complete-workflow samples.

For GPT-4o-mini, each task issues 32 sequential single-sample requests.
All 32 requests within a task use the same expansion prompt, and later
requests do not condition on earlier complete-workflow outputs. The
resulting complete outputs are deduplicated during candidate retention.

Qwen3.5-9B uses the corresponding 34-call CoT expansion configuration.
As with GPT-4o-mini, complete-workflow generations are sampled from the
same task-level expansion context rather than updated using previous
complete-workflow samples.

Thus, the realized CoT-tree configuration differs from IS primarily
in its generation scaffold and retention pathway, not through adaptive
tree-state exploration over complete workflows.

\paragraph{MO-Tree.}

MO-Tree uses five reasoning operators: rephrasing, direct answering,
Chain-of-Thought, divide-and-conquer, and self-refinement. Unlike the
realized CoT-tree configuration, later MO-Tree expansions may condition
on intermediate search states.

For GPT-4o-mini, MO-Tree uses two rollouts with nominal local sample
counts of up to 10 for direct answering, 16 for CoT, and 10 for
divide-and-conquer, with maximum search depth 5.

For Qwen3.5-9B, the budget-calibrated MO-Tree configuration uses one
rollout with local sample counts of 4 for direct answering, 8 for CoT,
and 5 for divide-and-conquer, with maximum search depth 5.

These MO-Tree configurations therefore differ from repeated complete
sampling in both operator composition and search-state dependence.
Their realized local branching behavior, however, depends on the
serving backend, as described below.

\subsubsection{Prompt-Matched Attribution Controls}
\label{app:prompt-matched}

For sensitivity analyses below, we refer to \emph{CoT-tree's retention
convention} as retaining complete JSON-array outputs while removing
duplicate raw completions. This is broader than text deduplication
alone because incomplete outputs are not retained.

\paragraph{GPT-4o-mini.}

To isolate the effect of the CoT-tree generation scaffold, we run a
prompt-matched IS control. For each of the 183 evaluation tasks, we
issue 32 independent single-sample requests using exactly the CoT-tree
root expansion prompt.

The prompt bytes are identical to those recorded for the corresponding
CoT-tree requests for all evaluation tasks. The control uses the same
decoding parameters as the primary GPT-4o-mini runs: temperature $0.7$,
top-$p=0.95$, maximum output length 2,048, and no request seed.

Across all 183 tasks, the control executes 5,856 successful requests.
Only one completion reaches the maximum output length. Its average
total inference cost is 242,193 tokens per task, compared with 242,476
for CoT-tree, a difference of approximately 0.1\%. The average
input-token count is identical across the two procedures
(226,527 tokens per task); the small total-cost difference comes from
completion length.

Applying CoT-tree's retention convention leaves both Oracle and
Selected counts unchanged for the GPT-4o-mini control. Its attribution
results and candidate-pool statistics are reported in
Section~\ref{sec:main-results} and Table~\ref{tab:app-pool-stats}.

\paragraph{Qwen3.5-9B.}

The Qwen prompt-matched control issues 34 independent single-sample
requests for each evaluation task using exactly the corresponding
CoT-tree expansion prompt. It uses temperature $0.7$, top-$p=0.95$,
a 2,048-token completion limit, no request seed, and thinking disabled.
Prompt hashes, prompt-token counts, and call counts match the CoT-tree
configuration task by task.

Across all 183 tasks, the control executes 6,222 successful requests;
38 completions (0.6\%) reach the maximum output length. Its average
inference cost is 252,666 tokens per task, compared with 252,678 and
252,445 in the two CoT-tree runs.

The control reaches 160 Oracle-correct tasks, compared with 162 and 157
for CoT-tree. The paired Oracle differences are $-2$ tasks
($p=.63$; bootstrap 95\% CI: $[-3.3,+1.1]$ pp) and $+3$ tasks
($p=.51$), respectively. Selected accuracy is 127, compared with 132
and 133 for CoT-tree; the paired differences are $-5$ tasks
($p=.23$; $[-6.6,+0.5]$ pp) and $-6$ tasks
($p=.15$; $[-7.1,+0.5]$ pp).

Applying CoT-tree's retention convention
(deduplication and removal of incomplete outputs) raises Selected
accuracy to 130, reducing these differences to $-2$ and $-3$ tasks
($p=.73$ and $.38$). Under this retention convention, the control
retains 11.24 candidates and 11.02 valid candidates per task, with
mean winning share .734 and 11 highest-vote ties.

The raw control averages 4.05 unique valid workflows per task,
compared with 3.91 and 4.01 for CoT-tree. Its mean winning share is
.797 under raw independent-sampling multiplicity and .734 under
CoT-tree's retention convention, compared with .733 and .740 for the
two CoT-tree runs. Invalidity is 1.4\% under the raw convention and
2.0\% under CoT-tree's retention convention, compared with 1.8\% and
2.5\% for the two CoT-tree runs. Thus, the control reproduces
CoT-tree-level candidate discovery and closely matches its diversity
at essentially identical realized cost.

\subsubsection{Run-Level Sensitivity of the Prompt-Matched Attribution}
\label{app:prompt-run-sensitivity}

Because candidate generation is stochastic, we additionally report the
run-level outcomes underlying the prompt-matched attribution comparison.
For Qwen3.5-9B, two independent CoT-tree runs are available under the
same decoding configuration. GPT-4o-mini has one primary CoT-tree run,
so its prompt-matched comparison should be interpreted as a matched
pipeline control rather than a variance estimate.

\begin{table*}[t]
    \centering
    \small
    \setlength{\tabcolsep}{5pt}
    \begin{tabular*}{\textwidth}{@{\extracolsep{\fill}}llrrrrr@{}}
        \toprule
        Model & Condition & Oracle & Selected & Tokens/task &
        Unique valid & Winner share \\
        \midrule

        GPT-4o-mini & CoT-tree
        & 153 & 131 & 242,476 & 4.22 & .769 \\

        GPT-4o-mini & Prompt-matched
        & 154 & 129 & 242,193 & 4.09 & .770 \\

        \midrule

        Qwen3.5-9B & CoT-tree R1
        & 162 & 132 & 252,678 & 3.91 & .733 \\

        Qwen3.5-9B & CoT-tree R2
        & 157 & 133 & 252,445 & 4.01 & .740 \\

        Qwen3.5-9B & Prompt-matched
        & 160 & 127 & 252,666 & 4.05 & .734 \\

        \bottomrule
    \end{tabular*}

    \caption{
    Run-level comparison for the prompt-matched attribution controls.
    Oracle and Selected are exact-match counts out of 183 tasks, and
    Tokens/task reports realized inference cost. Unique-valid workflow
    counts use the primary candidate pools. Winner share follows each
    procedure's retention convention; for the Qwen prompt-matched
    control, the reported .734 uses CoT-tree's retention convention.
    Qwen includes two independent CoT-tree runs under the same decoding
    configuration.
    }
    \label{tab:app-run-sensitivity}
\end{table*}

For attribution, Oracle coverage is the primary quantity because it
measures candidate discovery before final selection. The Qwen
prompt-matched control falls within the observed CoT-tree run range
(160 versus 162 and 157), while GPT-4o-mini differs by only one task
(154 versus 153). Selected accuracy is reported for completeness but is
affected by procedure-specific retention and voting semantics;
corresponding retention-sensitivity results are reported in
Appendix~\ref{app:prompt-matched}.

\subsection{Model and Decoding Parameters}
\label{app:decoding}

GPT-4o-mini is evaluated with temperature $0.7$, top-$p=0.95$, and a
maximum completion length of 2,048 tokens per generation call. All
GPT-4o-mini runs used in the primary comparisons, including the
prompt-matched control, omit the request seed.

Qwen3.5-9B uses temperature $0.7$, top-$p=0.95$, and the same
2,048-token completion limit. Individual Qwen generation requests do
not specify a request seed. Qwen is evaluated with thinking disabled
through the following template setting:

\begin{quote}
\small\ttfamily
chat\_template\_kwargs:\\
\{enable\_thinking: false\}
\end{quote}

Preliminary thinking-enabled experiments frequently exhausted the
completion budget before producing a complete workflow: in an earlier
thinking-enabled CoT run, 672 of 1,360 samples (49\%) produced no
workflow.

Qwen3.5-9B is served locally with vLLM through its OpenAI-compatible
interface. The server uses a maximum model length of 16,384 tokens,
prefix caching, the Qwen3 reasoning parser, and server-level seed 0.
Two independent Qwen generation runs are performed with the same
primary generation hyperparameters. Within each model, the seed policy
is held fixed across procedures, and all primary procedure comparisons
are made within model.

In both model settings, the standard completion limit is 2,048 tokens.
The only exception is the confidence tie-break used inside MO-Tree
local aggregation, which requests five short 10-token samples when two
candidate answers receive equal support.

We do not sweep temperature or top-$p$; these decoding parameters are
held fixed across the primary comparisons because the study focuses on
attribution under a common generation configuration rather than
decoding-parameter optimization.

\subsection{Backend-Realized Sampling Behavior}
\label{app:backend}

Nominal sampling parameters do not necessarily describe the behavior
realized by the serving backend. This distinction is especially
important for MO-Tree, whose search logic nominally permits multiple
samples at selected expansion nodes.

For GPT-4o-mini through OpenRouter, requests specifying $n>1$
nevertheless return exactly one completion. A direct diagnostic request
with $n=3$ returns one choice, and in the primary no-seed MO-Tree run
all 5,082 requests specifying $n>1$ likewise return a single
completion. Thus, nominal local multi-sample aggregation reduces to
single-sample updates for GPT-4o-mini. Consistently, the confidence
tie-break associated with local aggregation is never triggered.

In contrast, the local vLLM backend used for Qwen3.5-9B honors the
configured sample count. Across the two Qwen MO-Tree runs, all requests
with $n\in\{10,8,5,4,3\}$ return the requested number of completions:
4,607 such requests in Run~1 and 4,598 in Run~2. The local aggregation
logic is therefore active in both Qwen runs. The associated confidence
tie-break is triggered 216 times in Run~1 and 187 times in Run~2.

The GPT-4o-mini and Qwen MO-Tree experiments therefore share the same
nominal operator family and search logic but do not realize the same
local branching behavior. Cross-model MO-Tree differences cannot
therefore be attributed solely to model choice.

\subsection{Token and API-Call Accounting}
\label{app:resource-accounting}

Our compute comparison uses average aggregate inference-time tokens per
task rather than retained-candidate count or nominal branching width.
For each task, we sum prompt and completion tokens over all recorded
model calls contributing to the completed generation procedure.

A model call corresponds to one request to the serving backend. Thus,
a Qwen request with $n>1$ contributes one call even when it returns
multiple completions. This convention distinguishes backend request
count from the number of candidate completions realized within each
request.

The accounting includes generation, search expansion, and any
model-based local aggregation actually executed by the backend. A
generation that is later discarded, deduplicated, or absorbed during
candidate retention therefore still contributes to inference cost.
Offline normalization, structural validation, and deterministic
majority selection require no additional model inference.

The GPT-4o-mini configurations consume approximately 238k, 242k,
242k, and 239k tokens per task for IS, prompt-matched IS, CoT-tree,
and MO-Tree, respectively. For IS, CoT-tree, and MO-Tree, the exact
logged totals are 43,499,426, 44,373,195, and 43,806,026 tokens.
Detailed prompt-matched cost comparisons are reported in
Table~\ref{tab:app-run-sensitivity}.

For Qwen3.5-9B Run~1, the corresponding averages are approximately
263k, 253k, and 285k tokens per task for IS, CoT-tree, and MO-Tree.
Run~2 differs by less than 1k tokens per task for each procedure, with
corresponding averages of approximately 263k, 252k, and 284k.
The Qwen MO-Tree configuration is therefore described as
budget-calibrated rather than exactly compute matched.

\subsection{Exploratory Pooled Candidate Set}
\label{app:pooled-accounting}

An earlier exploratory pooled candidate set combines candidates from
two completed fixed-seed generation runs. It retains the first eight
candidates, ordered by \texttt{candidate\_id}, from an earlier
fixed-seed CoT-tree run and unions them with all candidates from a
separate narrower MO-Tree run.

The latter uses two rollouts, with local sample counts of 4 for CoT,
2 for direct answering, and 2 for divide-and-conquer, and maximum
depth 3.

A preliminary accounting estimate assigned cost proportionally to the
retained subset and yielded 231,860 tokens per task. This estimate was
incorrect because both contributing generation runs had already
incurred their full inference cost.

The complete realized cost is

\[
242{,}475 + 171{,}241 = 413{,}716
\]

tokens per task, or approximately $1.74\times$ the IS budget. We
therefore exclude this pooled set from all primary compute-controlled
comparisons and retain it only as a diagnostic example.

This case illustrates a general accounting principle: post-hoc
retention of a subset of generated candidates cannot remove the
inference cost required to produce the runs from which that subset was
obtained.

\subsection{Seed Sensitivity}
\label{app:seed}

The primary GPT-4o-mini comparisons omit request seeds. We additionally
examine fixed-seed runs as a sampling-sensitivity analysis. For IS,
the fixed-seed and no-seed runs were generated back to back with
unchanged generation code and configuration. Their per-task
prompt-token counts are identical, and the recorded request logs differ
in whether a request seed is supplied. This pair therefore provides
the cleanest seed-controlled comparison in our experiments.

The earlier CoT-tree and MO-Tree fixed-seed runs were not generated as
dedicated seed-only controls and are therefore reported only as broader
sensitivity diagnostics.

\begin{table*}[t]
    \centering
    \small
    \setlength{\tabcolsep}{4pt}
    \begin{tabular}{llrrrrr}
        \toprule
        Procedure & Sampling &
        Oracle &
        Selected &
        Retained/task &
        Unique valid/task &
        Winner share \\
        \midrule

        IS
        & fixed seed
        & 133 & 121 & 32.0 & 1.60 & .925 \\

        IS
        & no seed
        & 139 & 121 & 32.0 & 2.32 & .881 \\

        \midrule

        CoT-tree
        & fixed seed
        & 144 & 118 & 23.0 & 3.22 & .806 \\

        CoT-tree
        & no seed
        & 153 & 131 & 31.5 & 4.22 & .769 \\

        \midrule

        MO-Tree
        & fixed seed
        & 150 & 117 & 12.8 & 4.82 & .555 \\

        MO-Tree
        & no seed
        & 146 & 123 & 12.7 & 4.75 & .583 \\

        \bottomrule
    \end{tabular}

    \caption{
    GPT-4o-mini seed sensitivity.
    The primary runs omit request seeds.
    The IS fixed-seed row is a back-to-back matched rerun and provides
    the cleanest seed-controlled comparison.
    CoT-tree and MO-Tree use earlier fixed-seed runs and are therefore
    reported as broader sensitivity diagnostics rather than seed-only
    controls.
    Winner share is the mean vote share of the selected workflow.
    }
    \label{tab:seed-sensitivity}
\end{table*}

For IS, the matched fixed-seed and no-seed runs consume essentially
identical inference budgets. Removing the seed increases the mean
number of distinct raw responses from 2.03 to 3.96 and the mean number
of distinct valid workflows from 1.60 to 2.32. Winner concentration
correspondingly decreases from .925 to .881. The paired Oracle
difference is not statistically significant ($p=.11$), while Selected
accuracy remains unchanged at 121 tasks.

A larger difference appears between the earlier fixed-seed and primary
no-seed CoT-tree runs. For every evaluation task, its 32 generation
requests use the same prompt within that task. Under the fixed seed,
duplicate complete responses occur on 178 of 183 tasks, leaving 23.0
retained candidates per task after retention on average. Without the
seed, only 25 tasks contain any duplicate complete response, and the
retained pool increases to 31.5 candidates per task. Between these two
runs, Oracle rises from 144 to 153 tasks and Selected accuracy from 118
to 131. The paired differences are significant for both Oracle
($p=.023$) and Selected accuracy ($p=.004$). Because this pair was not
generated as a dedicated seed-only control, we treat it as a
sensitivity diagnostic rather than causal evidence for a seed effect.

MO-Tree realizes a different request pattern. In the no-seed run it
averages 30.5 requests and 30.5 distinct prompts per task because
prompts vary with search state or reasoning operator. Because its
prompts vary across requests, MO-Tree is not exposed to the
repeated-prompt duplication mechanism that arises when identical
prompts are repeatedly sampled with a fixed request seed. Oracle
changes from 150 to 146 tasks ($p=.42$), while Selected accuracy
changes from 117 to 123 ($p=.29$); neither difference is statistically
significant.

Taken together, the matched IS comparison shows that request seeding
can reduce candidate diversity under repeated identical-prompt
sampling, while the CoT-tree and MO-Tree comparisons provide broader
sensitivity evidence under different realized request patterns.

\section{Post-Generation Evaluation and Transformation Audit}
\label{app:transformation-audit}
\label{app:audit}

This appendix documents the common post-generation evaluation pipeline
and audits normalization choices that can affect workflow semantics,
structural validity, or downstream selection.

\subsection{Processing Pipeline and Recoverability}

Each candidate passes through JSON extraction, with automatic repair on
parse failure, followed by normalization, structural validation, and
selection. Oracle correctness, structural validation, and final
selection all operate on the normalized candidate.

One procedure-specific asymmetry was identified during the audit. The
original IS runs had passed through an additional generation-time
normalizer containing E2b and E4, whereas the CoT-tree and MO-Tree
runs had not. We therefore re-extracted every primary IS candidate
from its stored raw completion and reevaluated it using only the common
normalization pipeline. This changes the normalized representation of
17, 21, and 11 candidates for GPT-4o-mini, Qwen Run~1, and Qwen
Run~2, respectively, but changes the Exact Match status of no
candidate and leaves task-level Oracle and Selected results unchanged.
All reported IS results use these uniformly reprocessed candidates.
The prompt-matched IS controls were generated and evaluated directly
under the common protocol.

The purpose of normalization is to remove representation-level
variation without introducing new planning decisions. We call a
transformation \emph{semantically recoverable} when its normalized
value is uniquely determined by information already present in the
generated workflow together with the fixed tool interface. A
transformation that replaces or discards a model-generated planning
choice is not considered recoverable merely because a plausible
intended value can be inferred.

\subsection{Transformation Inventory and Targeted Ablations}

Table~\ref{tab:transformation-inventory} summarizes the transformations
relevant to the audit. Formatting-only operations and defensive code
paths that do not affect planning content are omitted.

\begin{table*}[t]
    \centering
    \small
    \setlength{\tabcolsep}{4pt}
    \begin{tabular}{llll}
        \toprule
        ID & Transformation & Recoverability status & Primary protocol \\
        \midrule

        E1
        & Generated-reference canonicalization
        & Recoverable
        & Enabled \\

        E2a
        & Dependence fields rebuilt from generated references
        & Recoverable
        & Enabled \\

        E2b
        & Multi-predecessor dependency reduction
        & Information-losing
        & Disabled \\

        E3
        & Empty outputs filled from the tool manifest
        & Recoverable
        & Enabled \\

        E4
        & Manifest-assisted output-key correction
        & Semantic correction
        & Disabled \\

        E5
        & Non-object inputs reset to an empty map
        & Not recoverable; never triggered
        & Disabled \\

        \bottomrule
    \end{tabular}

    \caption{
    Post-generation transformation inventory.
    E2b, E4, and E5 are disabled in the common protocol because they
    are not semantically recoverable under our criterion; E5 is never
    triggered in the evaluated candidate pools.
    }
    \label{tab:transformation-inventory}
\end{table*}

E1 canonicalizes generated-output references while preserving the same
source step and output key. E2a reconstructs
\texttt{dependence} and \texttt{dependence\_content} from generated
references in the step inputs while retaining all recovered
predecessors. E3 fills a missing or empty \texttt{outputs} list from
the invoked tool's declared outputs. None of E1--E3 changes the
selected tool, step order, step index, or a non-reference input value.

Across the nine primary pools, E1 affects 884 candidates, E2a affects
2,557, E3 affects 26, and E5 affects none. Across all 45,246
candidates in the eleven evaluated pools, normalization changes Exact
Match only from incorrect to correct: 1,612 candidates become exact
matches, while none changes in the reverse direction. This increases
the aggregate Oracle count from 1,649 to 1,686. These gains occur only
in non-IS pools. On GPT-4o-mini, normalization raises CoT-tree Oracle
from 145 to 153 and prompt-matched IS from 148 to 154 while leaving
IS unchanged at 139. Thus, the prompt-matched control matches or
exceeds CoT-tree under either convention. Removing E3 alone changes no
candidate-level Exact Match status or task-level result.

Normalization also rebuilds \texttt{dependence} before structural
validation. The validator's dependency--reference consistency checks
are therefore effectively redundant on the primary path: across all
45,246 candidates in the eleven evaluated pools, they fire on a single
candidate, which is also rejected for a malformed reference, and never
determine a rejection on their own.
\texttt{dependence\_content} is not checked by the validator.

\paragraph{E4: manifest-assisted output-key correction.}

E4 replaces an invalid generated output-key choice using information
from the tool manifest. Specifically, if a referenced output key is
not among those declared for the source step and that step declares
exactly one output, E4 substitutes the sole declared key.

Because this operation overwrites a model-generated planning choice,
we classify E4 as semantic correction and disable it in the common
evaluation. Enabling E4 changes some normalized representations and
secondary structural-validity statistics, but changes neither
candidate-level Exact Match nor task-level Oracle or Selected results
in the evaluated primary pools.

\paragraph{E2b: multi-predecessor dependency reduction.}

E2b was present in an earlier compatibility pipeline. When a
reconstructed dependency contains multiple generated predecessors, E2b
retains only the first. Because this operation discards generated
dependency information, we classify it as information-losing and
disable it in the common evaluation. No reference workflow contains a
step with more than one predecessor, so disabling E2b preserves all
dependency information expressed by candidate workflows without making
any reference workflow unrecoverable.

The fixed-seed pools used in the following E2b audit predate the
primary runs and are distinct from the matched fixed-seed IS rerun in
Appendix~\ref{app:seed}. In these earlier pools, E2b affects 195
candidate workflows but changes the Exact Match status of none. Its
effect is instead mediated through structural validity and the
resulting vote composition.

\begin{table}[t]
  \centering
  \scriptsize
  \renewcommand{\arraystretch}{1.1}
  \setlength{\tabcolsep}{2pt}
  \begin{tabular*}{\columnwidth}{@{\extracolsep{\fill}}lrrrr@{}}
    \toprule
    Method & \shortstack{Oracle\\E2b on} &
    \shortstack{Oracle\\E2b off} &
    \shortstack{Selected\\E2b on} &
    \shortstack{Selected\\E2b off} \\
    \midrule
    Fixed-seed IS & 132 & 132 & 118 & 117 \\
    CoT-tree       & 144 & 144 & 120 & 118 \\
    MO-Tree        & 150 & 150 & 117 & 117 \\
    \bottomrule
  \end{tabular*}

  \caption{
  Earlier fixed-seed GPT-4o-mini E2b compatibility audit.
  Disabling E2b leaves Oracle EM unchanged and affects Selected EM only
  through changes in structural validity and vote composition.
  }
  \label{tab:e2b-ablation}
\end{table}

The primary evaluation disables E2b for every pool. Re-enabling it
changes no primary Oracle result, changes GPT-4o-mini Selected accuracy
only for CoT-tree (131 to 132), and produces no task-level Oracle or
Selected changes for Qwen3.5-9B. E2b nevertheless changes some
candidate representations and structural-filtering decisions, so all
secondary statistics reported in the paper use the common
E2b-disabled evaluation.

\subsection{Normalization and Structural Filtering}

The original procedure-specific evaluation pipelines applied different
amounts of post-generation normalization. This created large apparent
differences in candidate invalidity and majority accuracy even when the
generated candidate pools were held fixed.

\begin{table}[t]
    \centering
    \small
    \setlength{\tabcolsep}{3pt}
    \begin{tabular}{lrrrr}
        \toprule
        & \multicolumn{2}{c}{Invalid (\%)} &
          \multicolumn{2}{c}{Unfiltered Majority} \\
        \cmidrule(lr){2-3}
        \cmidrule(lr){4-5}
        Method & Original & Common & Original & Common \\
        \midrule
        IS      & 2.0  & 0.9 & 117 & 117 \\
        CoT-tree & 27.4 & 2.9 & 104 & 118 \\
        MO-Tree  & 25.1 & 6.9 & 100 & 115 \\
        \bottomrule
    \end{tabular}

    \caption{
    Effect of common post-generation normalization on earlier
    fixed-seed GPT-4o-mini candidate pools.
    Invalidity is measured at the candidate level and majority values
    are correct-task counts out of 183.
    Generated candidate pools are held fixed, and the Common columns
    apply the common normalization protocol with E4 and E2b disabled.
    }
    \label{tab:app-normalization}
\end{table}

Under the original procedure-specific pipelines, candidate invalidity
is 2.0\% for IS, 27.4\% for CoT-tree, and 25.1\% for MO-Tree.
Applying the common normalization to the same generated pools reduces
these values to 0.9\%, 2.9\%, and 6.9\%, respectively. At the same
time, unfiltered-majority correctness changes from 117, 104, and 100
tasks to 117, 118, and 115.

This comparison shows that much of the originally apparent structural
invalidity gap arose from asymmetric upstream processing rather than
from intrinsic differences in the generated plans. Under the original
pipelines, structural filtering improves majority correctness by 11
tasks for CoT-tree and 9 for MO-Tree; under the common protocol applied
to the same earlier pools, the corresponding gains are 0 and 2 tasks.

In the primary GPT-4o-mini pools, structural filtering changes
majority correctness by 0, 2, and 1 tasks for IS, CoT-tree, and
MO-Tree, respectively; across the Qwen primary pools, the corresponding
change is at most one task. Across all eleven evaluated candidate
pools, structural filtering never removes the last correct workflow
from an Oracle-positive task. The validator therefore constrains
representation-level validity but does not resolve the semantic
tool-selection and dependency-construction errors that dominate the
remaining discovery--selection gap.

\paragraph{Filtering effects in the final evaluation.}
Across all eleven evaluated pools (2,013 pool--task cases),
structural filtering changes the selected workflow in 16
cases: six incorrect selections become correct, none of the
correct selections becomes incorrect, and ten remain
incorrect. Among 45,246 normalized candidates, no
exact-match candidate fails structural validation.
The fallback to the unfiltered pool is never triggered.
Consequently, filtering produces no filter losses in
the final DVS decomposition, although it can alter
majority-selection outcomes.

Per-transformation leave-one-out results for all evaluated pools are
included with the audit artifacts.

\section{Statistical Tests and Pool Diagnostics}
\label{app:additional-results}

This appendix reports paired statistical comparisons, calibration
sensitivity, and candidate-pool diagnostics supporting the main
results. Qwen3.5-9B generation runs are analyzed separately rather than
pooled as independent task observations.

\subsection{Paired Statistical Comparisons}
\label{app:stat-tests}

For each model setting, we report paired differences in correct-task
count, percentage-point difference, paired bootstrap 95\% confidence
intervals, and exact two-sided McNemar tests.

For GPT-4o-mini, the key attribution comparisons are
(1) prompt-matched IS versus the IS baseline and
(2) prompt-matched IS versus CoT-tree.
We also report CoT-tree--IS and MO-Tree--IS comparisons for
completeness.

\begin{table*}[t]
    \centering
    \small
    \setlength{\tabcolsep}{3.2pt}
    \begin{tabular}{llrrrr}
        \toprule
        Model / Run & Comparison &
        Metric &
        $\Delta$ tasks &
        $\Delta$ pp [95\% CI] &
        McNemar $p$ \\
        \midrule

        GPT-4o-mini
        & Prompt-IS $-$ IS
        & Oracle
        & +15
        & +8.2 [+3.8, +13.1]
        & .0015 \\

        GPT-4o-mini
        & Prompt-IS $-$ CoT-tree
        & Oracle
        & +1
        & +0.5 [-3.3, +4.4]
        & 1.000 \\

        GPT-4o-mini
        & CoT-tree $-$ IS
        & Oracle
        & +14
        & +7.7 [+3.3, +12.6]
        & .0026 \\

        GPT-4o-mini
        & MO-Tree $-$ IS
        & Oracle
        & +7
        & +3.8 [-1.1, +8.7]
        & .210 \\

        \midrule

        GPT-4o-mini
        & Prompt-IS $-$ IS
        & Selected
        & +8
        & +4.4 [-1.1, +9.8]
        & .169 \\

        GPT-4o-mini
        & Prompt-IS $-$ CoT-tree
        & Selected
        & -2
        & -1.1 [-4.4, +1.6]
        & .727 \\

        GPT-4o-mini
        & CoT-tree $-$ IS
        & Selected
        & +10
        & +5.5 [0.0, +11.5]
        & .087 \\

        GPT-4o-mini
        & MO-Tree $-$ IS
        & Selected
        & +2
        & +1.1 [-4.9, +7.1]
        & .860 \\

        \midrule

        Qwen Control
        & Prompt-IS $-$ CoT-tree Run 1
        & Oracle
        & -2
        & -1.1 [-3.3, +1.1]
        & .630 \\

        Qwen Control
        & Prompt-IS $-$ CoT-tree Run 2
        & Oracle
        & +3
        & +1.6 [-1.6, +4.9]
        & .508 \\

        Qwen Control
        & Prompt-IS $-$ CoT-tree Run 1
        & Selected
        & -5
        & -2.7 [-6.6, +0.5]
        & .230 \\

        Qwen Control
        & Prompt-IS $-$ CoT-tree Run 2
        & Selected
        & -6
        & -3.3 [-7.1, +0.5]
        & .150 \\

        \midrule

        Qwen Run 1
        & CoT-tree $-$ IS
        & Oracle
        & +3
        & +1.6 [-1.6, +5.5]
        & .549 \\

        Qwen Run 1
        & MO-Tree $-$ IS
        & Oracle
        & -12
        & -6.6 [-10.9, -2.2]
        & .008 \\

        Qwen Run 1
        & CoT-tree $-$ IS
        & Selected
        & +3
        & +1.6 [-2.2, +5.5]
        & .581 \\

        Qwen Run 1
        & MO-Tree $-$ IS
        & Selected
        & -7
        & -3.8 [-8.7, +1.1]
        & .210 \\

        \midrule

        Qwen Run 2
        & CoT-tree $-$ IS
        & Oracle
        & -5
        & -2.7 [-7.1, +1.1]
        & .302 \\

        Qwen Run 2
        & MO-Tree $-$ IS
        & Oracle
        & -15
        & -8.2 [-13.1, -3.8]
        & .001 \\

        Qwen Run 2
        & CoT-tree $-$ IS
        & Selected
        & +5
        & +2.7 [-0.5, +6.6]
        & .227 \\

        Qwen Run 2
        & MO-Tree $-$ IS
        & Selected
        & -6
        & -3.3 [-7.7, +1.1]
        & .238 \\

        \bottomrule
    \end{tabular}

    \caption{
    Paired statistical comparisons.
    Prompt-IS denotes prompt-matched IS.
    Confidence intervals are obtained by paired bootstrap over tasks;
    $p$-values are from exact two-sided McNemar tests.
    }
    \label{tab:app-paired-tests}
\end{table*}

On GPT-4o-mini, prompt-matched IS exceeds the IS baseline by 15
Oracle-correct tasks ($+8.2$ percentage points; $p=.0015$), closely
matching the CoT-tree--IS difference of 14 tasks. Prompt-matched IS
and CoT-tree differ by only one Oracle task and two Selected tasks;
neither difference is statistically significant (Oracle $p=1.000$;
Selected $p=.727$), and both bootstrap intervals include zero.

The corresponding Selected improvements over IS are smaller:
prompt-matched IS gains eight tasks ($+4.4$ points, $p=.169$), while
CoT-tree gains ten ($+5.5$ points, $p=.087$). Thus, the discovery gain
reproduced by prompt-matched sampling does not establish a
corresponding statistically reliable improvement in final selection.
MO-Tree differs significantly from IS on neither GPT-4o-mini metric.

On Qwen3.5-9B, prompt-matched sampling lies within the variation of the
two CoT-tree runs and does not differ significantly from either.
CoT-tree is also statistically indistinguishable from IS on both
Oracle and Selected EM in both runs. The budget-calibrated MO-Tree has
significantly lower Oracle coverage than IS in both runs, while its
Selected differences are not statistically significant.

Taken together, the prompt-matched controls reproduce CoT-tree-level
discovery on both models, and changes in discovery do not translate
proportionally into final selection.

\subsection{Calibration Exclusion}
\label{app:calibration-exclusion}

Five GPT-4o-mini evaluation tasks were used during search-budget
calibration. We therefore repeat the IS, CoT-tree, and MO-Tree
comparisons after excluding these tasks, leaving 178 evaluation tasks.

\begin{table}[t]
    \centering
    \small
    \setlength{\tabcolsep}{3.5pt}
    \begin{tabular}{lrrrr}
        \toprule
        Procedure &
        Oracle &
        Selected &
        $\Delta$ Oracle &
        $\Delta$ Selected \\
        \midrule
        IS      & 135 & 117 & -- & -- \\
        CoT-tree & 148 & 127 & +13 (.004) & +10 (.087) \\
        MO-Tree  & 142 & 119 & +7 (.210)  & +2 (.860) \\
        \bottomrule
    \end{tabular}

    \caption{
    GPT-4o-mini sensitivity analysis after excluding the five
    evaluation tasks used during search-budget calibration ($n=178$).
    Differences are measured relative to the primary no-seed IS
    baseline; exact two-sided McNemar $p$-values are in parentheses.
    }
    \label{tab:app-calibration-exclusion}
\end{table}

On the five excluded tasks, IS, CoT-tree, and MO-Tree are
Oracle-correct on 4, 5, and 4 tasks, respectively, while all three are
Selected-correct on four tasks. After exclusion, the CoT-tree--IS
difference remains similar to that observed on all 183 tasks.

This sensitivity analysis does not alter the attribution result: the
full 183-task prompt-matched control already shows that the apparent
CoT-tree Oracle advantage is reproduced by independent sampling with
the same generation scaffold. The calibration-exclusion experiment
should therefore be interpreted as a robustness check on the primary
comparison rather than as evidence for a tree-structure effect.

The paired bootstrap intervals for the 178-task comparison are also
similar to those in the full analysis: CoT-tree yields
$[+2.8,+11.8]$ pp for Oracle and $[0.0,+11.2]$ pp for Selected,
whereas MO-Tree yields $[-1.1,+9.6]$ pp and $[-5.1,+7.3]$ pp,
respectively. 
\paragraph{Prompt-matched attribution control.}
Excluding the five calibration tasks preserves the main
attribution finding. On the remaining 178 tasks, prompt-matched
IS achieves 150 Oracle-correct workflows, compared with 148
for CoT-tree (+1.1 percentage points; $p=.754$;
paired bootstrap 95\% CI: $[-2.2,+4.5]$ pp).
Selected accuracy is 125 versus 127 ($-1.1$ percentage
points; $p=.727$; 95\% CI: $[-4.5,+1.7]$ pp).
Neither paired difference is statistically significant.
Thus, the prompt-matched control continues to reproduce
CoT-tree-level discovery after excluding tasks used for
budget calibration.

\subsection{Candidate-Pool Statistics}
\label{app:pool-stats}

Candidate pools differ substantially in retention, diversity, and vote
concentration because of their generation and retention procedures.
Table~\ref{tab:app-pool-stats} summarizes descriptive statistics after
common post-generation normalization, using each procedure's primary
retention convention.

\begin{table*}[t]
    \centering
    \small
    \setlength{\tabcolsep}{3.0pt}
    \begin{tabular}{llrrrrr}
        \toprule
        Model / Run & Procedure &
        Retained &
        Valid &
        Unique valid WF &
        Winning share &
        Ties \\
        \midrule

        GPT-4o-mini
        & IS
        & 32.0 & 31.7 & 2.32 & .881 & 1 \\

        GPT-4o-mini
        & Prompt-matched IS
        & 32.0 & 31.1 & 4.09 & .770 & 3 \\

        GPT-4o-mini
        & CoT-tree
        & 31.5 & 30.6 & 4.22 & .769 & 5 \\

        GPT-4o-mini
        & MO-Tree
        & 12.7 & 11.9 & 4.75 & .583 & 17 \\

        \midrule

        Qwen Control
        & Prompt-matched IS
        & 34.0 & 33.5 & 4.05 & .797 & 5 \\

        \midrule

        Qwen Run 1
        & IS
        & 34.0 & 28.7 & 4.29 & .776 & 6 \\

        Qwen Run 1
        & CoT-tree
        & 11.2 & 11.0 & 3.91 & .733 & 14 \\

        Qwen Run 1
        & MO-Tree
        & 7.3 & 6.2 & 2.67 & .689 & 20 \\

        \midrule

        Qwen Run 2
        & IS
        & 34.0 & 28.3 & 4.31 & .773 & 4 \\

        Qwen Run 2
        & CoT-tree
        & 11.1 & 10.8 & 4.01 & .740 & 8 \\

        Qwen Run 2
        & MO-Tree
        & 7.5 & 6.2 & 2.48 & .729 & 17 \\

        \bottomrule
    \end{tabular}

    \caption{
    Candidate-pool statistics after common post-generation
    normalization, averaged per task except Ties, which counts tasks
    with a highest-vote tie. Statistics follow each procedure's primary
    retention convention. Winning share is the fraction of valid
    retained candidate multiplicity assigned to the selected workflow
    group. Prompt-matched rows use raw independent-sampling
    multiplicity; under CoT-tree's retention convention, the Qwen
    control retains 11.2 candidates per task with winning share .734
    (Appendix~\ref{app:prompt-matched}).
    }
    \label{tab:app-pool-stats}
\end{table*}

The candidate-pool statistics reinforce the GPT attribution result.
Relative to IS, prompt-matched IS increases the mean number of unique
valid workflows from 2.32 to 4.09 and reduces mean winning
concentration from .881 to .770. These values closely match CoT-tree
at 4.22 unique valid workflows and .769 winning share.

Among prompt-matched IS's 24 strict vote-loss cases, the winning
incorrect workflow receives 48.6\% of valid votes on average
(median 42.7\%), compared with 67.8\% for IS and 41.3\% for
CoT-tree. These vote-loss-specific concentrations are distinct from
the overall mean winning shares in Table~\ref{tab:app-pool-stats}, and
again place the prompt-matched pool closer to CoT-tree than to IS.

Thus, the prompt-matched control reproduces not only CoT-tree's Oracle
coverage but also key diversity and concentration characteristics of
its realized candidate pool. These similarities make retention
differences unlikely to explain the GPT attribution result.

The GPT-4o-mini MO-Tree pool retains fewer candidates but produces the
largest mean number of unique valid workflows among the three primary
procedures. Its lower winning share (.583) and larger number of ties
describe a more dispersed retained pool; these descriptive statistics
should not be interpreted as causing its accuracy differences.

The Qwen pools likewise show procedure-dependent retention, diversity,
and concentration. Across both runs, IS retains substantially more
valid candidates, CoT-tree retains an intermediate-sized pool, and
MO-Tree retains the smallest pool. Because retention, aggregation,
prompting, and backend behavior jointly determine realized candidate
multiplicity, these descriptive statistics do not by themselves
identify the cause of performance differences.

\section{Generation-Budget and Search-Configuration Sensitivity}
\label{app:budget-sensitivity}

This appendix provides additional analyses of how candidate discovery
and selection change with generation budget and MO-Tree search width.
The purpose is to separate the effect of generating more candidates
from the behavior of particular generation configurations.

\subsection{Generation-Budget Scaling}
\label{app:budget-prefix}

For IS and the realized CoT-tree configuration, complete-workflow
generation is non-adaptive:
later calls do not alter outputs produced by earlier calls.
We therefore reconstruct chronological budget prefixes by retaining the
first $N$ generation calls.

IS issues its calls sequentially, with candidate identifiers
corresponding to generation order.
CoT-tree creates complete-workflow leaves through a sequential loop of
single-sample calls, and the stored leaf order matches generation
order.
Every task contains exactly 32 GPT-4o-mini leaves or 34 Qwen leaves,
so no generation call is missing from a reconstructed prefix.

Within each task, all CoT-tree complete-workflow calls use the same
expansion prompt, and later calls do not condition on earlier generated
workflows.
Thus, the prefix analysis for this configuration should be interpreted
as repeated generation under the CoT-tree scaffold rather than as
progressive adaptive tree exploration.
At the full budget, the reconstructed prefixes exactly reproduce the
corresponding final evaluation results.

MO-Tree is adaptive:
later search states may depend on earlier expansions and retained
nodes.
A post-hoc prefix of a completed run is therefore not equivalent to
rerunning MO-Tree with a smaller search budget.
For this reason, MO-Tree is shown only at complete-run operating
points.

\begin{figure*}[t]
    \centering
    \includegraphics[width=\textwidth]{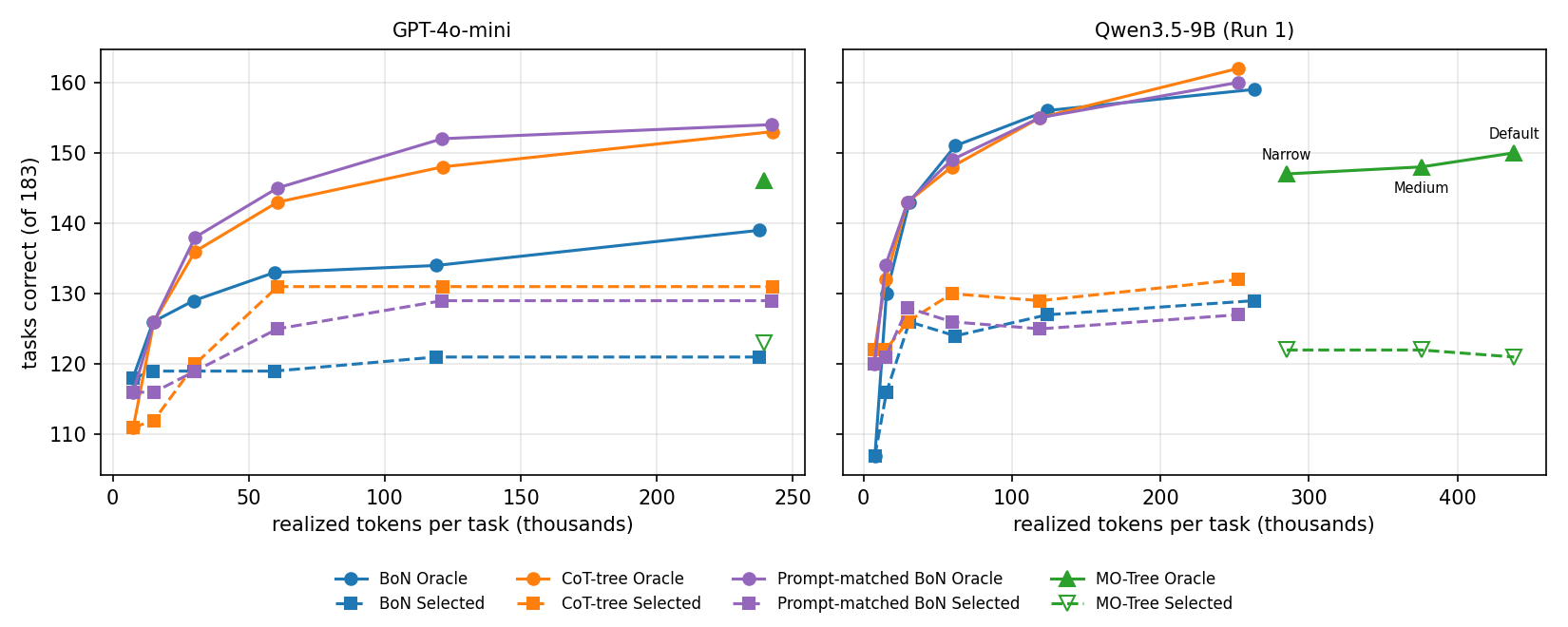}
    \caption{
    Generation-budget scaling for GPT-4o-mini and Qwen3.5-9B Run~1.
    For IS and CoT-tree, chronological prefixes of
    $N=1,2,4,8,16$ calls and the full budget are evaluated.
    Because candidate-level token usage is not available for every
    intermediate call, prefix locations on the horizontal axis are
    approximated as $N/N_{\mathrm{full}}$ of the full-run realized
    token count.
    CoT-tree prefixes correspond to repeated independent
    complete-workflow generations under the same expansion scaffold,
    not adaptive tree-state updates.
    MO-Tree is shown only at complete-run configurations because its
    adaptive state does not admit equivalent post-hoc prefixes.
    }
    \label{fig:app-budget-scaling}
\end{figure*}

Figure~\ref{fig:app-budget-scaling} reports Oracle and Selected EM
across generation budgets.

The GPT-4o-mini curves make the discovery--selection separation
especially clear.
For the realized CoT-tree configuration, Selected accuracy reaches its
final value of 131 correct tasks by eight generation calls and remains
unchanged at 16 and 32 calls, whereas Oracle coverage continues to
increase to 153 tasks.
The original IS curve ends at 139 Oracle-correct and 121
Selected-correct tasks; its Selected accuracy increases only from
118 after one call to 121 at the full budget.

These curves should not be interpreted as evidence for a tree-structure
effect. Section~\ref{sec:main-results} shows that independent sampling
with the same CoT-tree expansion prompt reproduces the same discovery
advantage over IS at the full budget. Across the reconstructed
GPT-4o-mini prefixes, prompt-matched IS reproduces this discovery
advantage over IS from four calls onward. The budget analysis instead
shows that, within a fixed generation scaffold, additional samples can
continue increasing correct-workflow coverage after Selected accuracy
has largely saturated.

The same qualitative pattern appears for Qwen3.5-9B.
For example, Qwen Run~1 IS increases from 107 Oracle-correct tasks
after one call to 159 at the full 34-call budget, whereas Selected
accuracy remains between 124 and 129 tasks from four calls onward.
Thus, additional generation can continue improving the availability of
correct workflows after downstream selection has largely saturated.

Because per-call token usage is not linked to individual candidates in
every run, intermediate token positions are approximate.
The token axis is nevertheless useful for comparing complete operating
points across procedures, particularly for Qwen MO-Tree, where one
backend request can return multiple completions.
The chronological prefix analysis itself is indexed by generation
calls; plotted token positions are a linear rescaling of prefix size.

\subsubsection{Random Candidate-Subsampling Sensitivity}
\label{app:random-subsampling}

As a complementary diagnostic, we subsample completed candidate pools
rather than reconstruct chronological generation prefixes.
For each candidate count, we draw 50 subsets without replacement and
report mean Oracle and Selected accuracy.

Sampling operates over raw generation calls before normalization,
structural filtering, or selection.
A Python random generator with seed 0 is re-initialized separately for
each pool and candidate count.
For CoT-tree, duplicate candidates are removed after the subset is
drawn, following its primary candidate-processing procedure.
The full-budget point is evaluated once rather than repeatedly
subsampled.

\begin{figure*}[t]
    \centering
    \includegraphics[width=\textwidth]{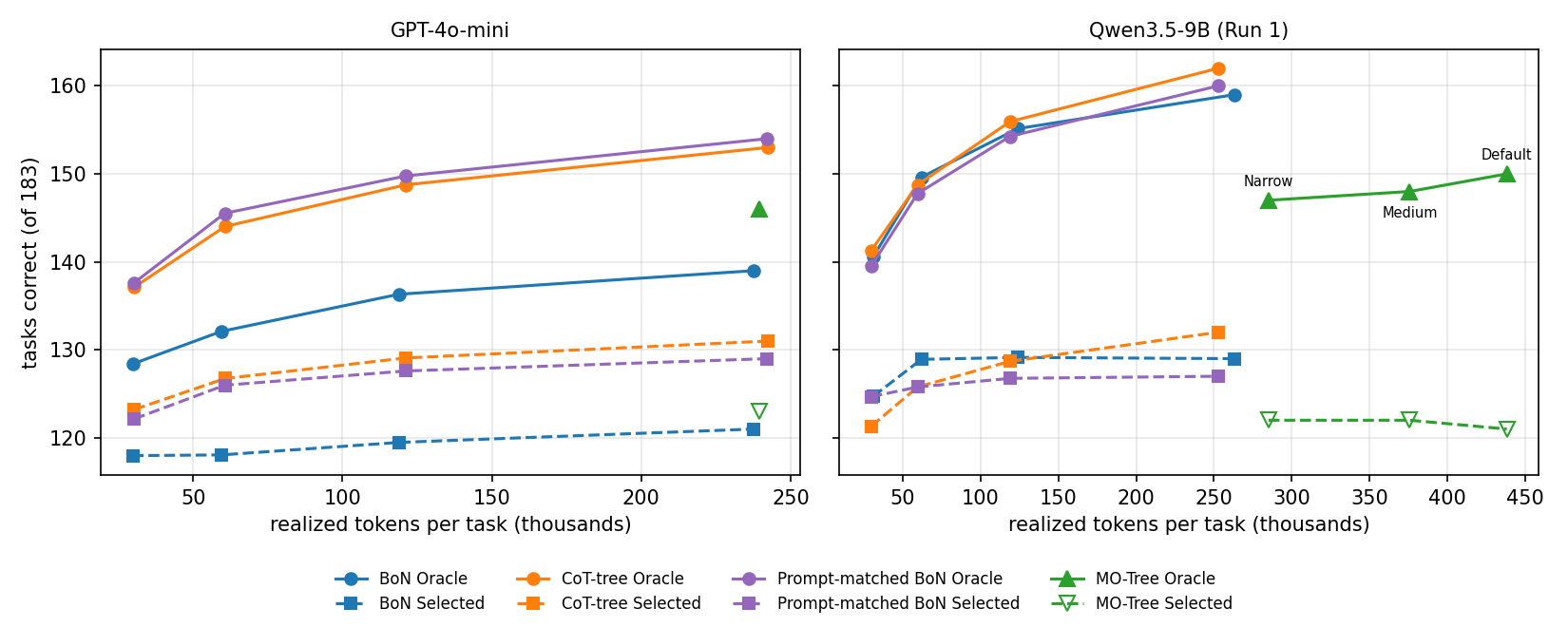}
    \caption{
    Candidate-budget sensitivity under random subsampling.
    Each intermediate point averages 50 subsets drawn without
    replacement from the completed raw generation pool.
    This analysis measures sensitivity to candidate budget rather than
    the behavior of an actual lower-budget adaptive search procedure.
    MO-Tree points correspond to complete-run width configurations.
    }
    \label{fig:app-candidate-subsampling}
\end{figure*}

This diagnostic produces the same qualitative separation:
candidate coverage generally increases more consistently than final
Selected accuracy as larger subsets are retained.
Recomputing the Qwen curves under the final E2b-disabled evaluation
leaves Oracle means unchanged and changes Selected means by at most
0.1 tasks.
We therefore use chronological prefixes for the main budget analysis
and retain random subsampling as a robustness diagnostic.

\subsection{Qwen MO-Tree Width Sensitivity}
\label{app:qwen-width}

The primary Qwen MO-Tree configuration was narrowed during budget
calibration.
To test whether its lower Oracle coverage is mainly a consequence of
this width reduction, we additionally evaluate medium and
original-width configurations while holding the model, prompting, and
downstream evaluation protocol fixed.

\begin{table*}[t]
    \centering
    \small
    \setlength{\tabcolsep}{4pt}
    \begin{tabular}{lrrrrrrr}
        \toprule
        MO width &
        Tokens/task &
        Calls/task &
        Retained &
        Valid &
        Unique WF &
        Oracle &
        Selected \\
        \midrule
        Narrow   & 285k & 33.6 & 7.3  & 6.2 & 2.67 & 147 & 122 \\
        Medium   & 376k & 42.7 & 10.1 & 8.6 & 2.90 & 148 & 122 \\
        Original & 438k & 48.6 & 11.7 & 9.8 & 2.95 & 150 & 121 \\
        \bottomrule
    \end{tabular}

    \caption{
    Qwen3.5-9B Run~1 MO-Tree width sensitivity under the
    E2b-disabled evaluation.
    Wider search substantially increases realized compute and retained
    candidates but yields only small gains in distinct-workflow
    coverage and Oracle accuracy, with no Selected improvement.
    These configurations are diagnostic rather than compute-matched
    comparisons against IS or CoT-tree.
    }
    \label{tab:app-qwen-width}
\end{table*}

Increasing the configuration from narrow to original width raises
average token usage from approximately 285k to 438k tokens per task,
an increase of about 54\%.
The mean number of valid retained candidates increases from 6.2 to
9.8, but the mean number of distinct valid workflows increases only
from 2.67 to 2.95.
Oracle improves by three tasks, from 147 to 150, while Selected
decreases from 122 to 121.

The Oracle increase from narrow to original width is not statistically
significant ($p=.61$).
The medium-width configuration differs from narrow by only one
Oracle-correct task ($p=1.00$).

Even the original-width MO-Tree remains below Qwen Run~1 IS by nine
Oracle-correct tasks ($p=.035$) and below the realized CoT-tree
configuration by 12 tasks ($p=.008$), despite using substantially more
inference compute.
We therefore find little evidence that the lower Qwen MO-Tree coverage
is primarily an artifact of the narrower budget-calibrated
configuration.

In this setting, substantially greater MO-Tree width and inference
expenditure are accompanied by only a small increase in the number of
distinct valid workflows.

\subsection{Active Tree-Selection Pilot}
\label{app:active-uct}

We conduct an additional 40-task pilot to examine whether
activating reward-guided UCT selection improves workflow
discovery or final selection. We compare independent sampling
(IS), MO-Tree with random tree selection, and MO-Tree with
active UCT selection. The two MO-Tree conditions share the
same generation configuration and differ only in their
tree-selection policy.

\begin{table}[t]
\centering
\small
\setlength{\tabcolsep}{2pt}
\begin{tabular*}{\columnwidth}{@{\extracolsep{\fill}}lrrrr@{}}
\toprule
Procedure & Oracle & Selected & Gap & Tok./task \\
\midrule
IS + majority & 35 & 32 & 3 & 989k \\
MO-Tree + Random & 37 & 29 & 8 & 1,064k \\
MO-Tree + Active UCT & 35 & 31 & 4 & 979k \\
\bottomrule
\end{tabular*}
\caption{
Active tree-selection pilot on 40 tasks.
Results use the common DisasterBench evaluation protocol.
Inference costs are realized tokens per task.
No paired Oracle or Selected comparison is statistically
significant under exact two-sided McNemar tests.
}
\label{tab:active-uct-pilot}
\end{table}

\paragraph{Performance comparison.}
Table~\ref{tab:active-uct-pilot} reports results under
approximately matched inference budgets. Active UCT
achieves 35 Oracle-correct and 31 Selected-correct tasks,
compared with 37/29 for random tree selection and 35/32
for IS. Average token expenditure is 979k, 1,064k,
and 989k per task, respectively. None of the paired
differences is statistically significant under exact
two-sided McNemar tests. Thus, this pilot provides no
evidence of an accuracy benefit from active UCT in
the evaluated configuration.

\paragraph{Realized UCT behavior.}
Execution-trace replay confirms that active UCT performs
501 nontrivial tree-selection decisions, each involving
at least three available child nodes. In 499 decisions,
the children have different mean rewards. Active UCT
selects the highest-scoring child under its current
visit counts and action values in 457 cases (91.2\%),
with the remaining decisions reflecting exploration.
Removing the value term would change the selected child
in 255 decisions (50.9\%). Moreover, UCT selects the
highest-mean-reward child in 248 cases, compared with
146 expected under uniform random selection. These
results confirm that the implemented policy uses reward
information and changes search allocation.

\begin{table}[t]
\centering
\small
\setlength{\tabcolsep}{4pt}
\begin{tabular}{lrr}
\toprule
Rollout outcome & Active UCT & Random \\
\midrule
No new nodes & 210 (32.8\%) & 156 (24.4\%) \\
Terminal CoT leaf revisits & 197 (30.8\%) & 134 (20.9\%) \\
New retained candidates & 310 (48.4\%) & 326 (50.9\%) \\
\midrule
Retained candidates/task & 17.8 & 18.7 \\
\bottomrule
\end{tabular}
\caption{
Realized search behavior in the 40-task pilot
(640 rollouts per tree-selection policy).
Terminal-leaf revisits repeat an already reached CoT
solution node without invoking the LLM.
A rollout is counted as producing new retained candidates
when at least one candidate in the final deduplicated pool
is attributed to that rollout. The first three rows are
not mutually exclusive.
}
\label{tab:active-uct-behavior}
\end{table}

\paragraph{Interpretation.}
Although active UCT uses reward-sensitive tree selection,
it provides no statistically significant improvement in
Oracle or Selected accuracy in this pilot. Its allocation
is also less productive under the implemented search
structure: 210 of 640 UCT rollouts (32.8\%) visit no new
nodes, compared with 156 (24.4\%) under random selection
(Table~\ref{tab:active-uct-behavior}). UCT also revisits
terminal CoT leaves more frequently (197 versus 134
rollouts). Such revisits repeat reward backpropagation
without generating additional workflows.

Conversely, only 310 UCT rollouts are associated with
new retained workflow candidates, compared with 326
under random selection, and UCT retains fewer distinct
candidates per task (17.8 versus 18.7). This suggests a
specific implementation-level limitation: reward-guided
selection can favor previously evaluated terminal
solutions without converting additional search effort
into candidate discovery. The 40-task sample and high
baseline Oracle coverage preclude conclusions about
reward-guided tree search more generally.

\section{Selection and Vote-Loss Analysis}
\label{app:selection-analysis}

This appendix provides detailed analyses of workflow rank, semantic
vote-loss errors, recurrence, selector sensitivity, and targeted
verification supporting Section~\ref{sec:analysis}.

The two prompt-matched IS controls are included in the stage-wise,
rank-based, diversity, and vote-margin analyses, but are not treated as
additional primary procedures in the selector, verifier, or 60-case
semantic-audit experiments. Unless otherwise noted, those analyses use
the nine primary candidate pools.

\subsection{Rank and Vote-Margin Diagnostics}
\label{app:topk-selection}

We first examine where the best correct workflow lies in the vote
ranking after structural filtering. Normalized valid candidates are
grouped by exact workflow identity, groups are ordered by descending
vote count, and vote ties are broken by the smallest candidate ID in
the group. This exactly matches the primary selector; as a sanity
check, Top-1 reproduces Selected EM in all eleven evaluated pools.

For strict vote-loss cases, we record the rank of the highest-ranked
exact-match workflow. Top-$k$ availability is slightly broader: for every task,
it counts whether an exact-match workflow appears among
the first $k$ ranked workflow groups. It
therefore also includes tie-loss cases in which the correct workflow
has the same vote count as the selected group but loses deterministic
tie-breaking.

\begin{table*}[t]
    \centering
    \small
    \setlength{\tabcolsep}{3.2pt}
    \begin{tabular}{llrrrrrrrr}
        \toprule
        Model / Run & Procedure &
        Vote loss & Rank 2 & Rank 3 & $>3$ &
        Selected & Top-2 & Top-3 & Oracle \\
        \midrule

        GPT-4o-mini
        & IS
        & 18 & 11 & 5 & 2
        & 121 & 132 & 137 & 139 \\

        & CoT-tree
        & 22 & 10 & 2 & 10
        & 131 & 141 & 143 & 153 \\

        & MO-Tree
        & 21 & 13 & 3 & 5
        & 123 & 138 & 141 & 146 \\

        & Prompt-matched IS$^\dagger$
        & 24 & 13 & 6 & 5
        & 129 & 143 & 149 & 154 \\

        \midrule

        Qwen3.5-9B Control
        & Prompt-matched IS$^\dagger$
        & 32 & 17 & 7 & 8
        & 127 & 145 & 152 & 160 \\

        \midrule

        Qwen3.5-9B Run 1
        & IS
        & 28 & 17 & 7 & 4
        & 129 & 148 & 155 & 159 \\

        & CoT-tree
        & 28 & 14 & 5 & 9
        & 132 & 148 & 153 & 162 \\

        & MO-Tree
        & 19 & 15 & 4 & 0
        & 122 & 141 & 146 & 147 \\

        \midrule

        Qwen3.5-9B Run 2
        & IS
        & 33 & 20 & 8 & 5
        & 128 & 149 & 157 & 162 \\

        & CoT-tree
        & 22 & 12 & 3 & 7
        & 133 & 147 & 150 & 157 \\

        & MO-Tree
        & 18 & 9 & 5 & 4
        & 122 & 134 & 140 & 147 \\

        \bottomrule
    \end{tabular}

    \caption{
    Rank of the highest-ranked correct workflow in strict vote-loss
    cases and Top-$k$ oracle availability. Rank columns are defined
    only over strict vote losses, whereas Top-$k$ columns are evaluated
    over all tasks and therefore also include tie-loss cases.
    Top-$k$ indicates whether a correct workflow is available within the
    first $k$ ranked groups, not an achieved correction rate.
    $^\dagger$Prompt-matched IS rows are attribution controls rather
    than additional primary procedures.
    }
    \label{tab:app-topk-selection}
\end{table*}

Across the nine primary pools, there are 209 strict vote losses. The
highest-ranked correct workflow is second in 121 cases (57.9\%),
third in 42 (20.1\%), and below third in 46 (22.0\%). The same pools
contain 231 Oracle--Selected losses in total; Top-2 makes a correct
workflow available within the top two groups for 137 of these losses
(59.3\%), while Top-3 does so for 181 (78.4\%).

Including both prompt-matched attribution controls yields 265 strict
vote losses and 289 Oracle--Selected losses across all eleven pools.
The highest-ranked correct workflow is second in 151 cases (57.0\%),
third in 55 (20.8\%), and below third in 59 (22.3\%). Top-2 makes a
correct workflow available for 169 Oracle--Selected losses (58.5\%),
while Top-3 does so for 226 (78.2\%).

Vote-loss tasks also contain more distinct valid workflow groups than
typical tasks. Across all eleven evaluated pools, the median is five
distinct valid groups on vote-loss tasks versus two across all tasks,
and the same direction holds in every individual pool. Thus, the high
rank of correct workflows is not merely a consequence of universally
tiny candidate pools.

\paragraph{Vote margins.}

\begin{table*}[t]
    \centering
    \small
    \setlength{\tabcolsep}{3.2pt}
    \begin{tabular}{llrrrrr}
        \toprule
        Model / Run & Procedure &
        \shortstack{Correct share\\mean} &
        \shortstack{Vote deficit\\mean} &
        \shortstack{Rank-2 deficit\\median} &
        $\leq 1$ vote & $\leq 2$ votes \\
        \midrule

        GPT-4o-mini
        & IS
        & .154 & 16.3 & 18 & 0 & 1 \\

        & CoT-tree
        & .147 & 8.1 & 5 & 2 & 4 \\

        & MO-Tree
        & .219 & 2.2 & 2 & 7 & 15 \\

        & Prompt-matched IS$^\dagger$
        & .162 & 10.0 & 7 & 1 & 4 \\

        \midrule

        Qwen3.5-9B Control
        & Prompt-matched IS$^\dagger$
        & .138 & 13.3 & 16 & 1 & 5 \\

        \midrule

        Qwen3.5-9B Run 1
        & IS
        & .169 & 11.0 & 10 & 1 & 1 \\

        & CoT-tree
        & .144 & 3.6 & 3.5 & 4 & 9 \\

        & MO-Tree
        & .229 & 2.3 & 2 & 9 & 14 \\

        \midrule

        Qwen3.5-9B Run 2
        & IS
        & .176 & 10.1 & 8.5 & 2 & 5 \\

        & CoT-tree
        & .166 & 3.6 & 3 & 6 & 10 \\

        & MO-Tree
        & .212 & 2.6 & 3 & 6 & 9 \\

        \bottomrule
    \end{tabular}

    \caption{
    Vote-margin diagnostics for strict vote-loss cases. Correct share
    is the mean vote share of the highest-ranked correct workflow;
    vote deficit is the selected wrong group's vote count minus that of
    the best correct group. Rank-2 deficit is reported only when the
    best correct workflow ranks second. Absolute deficits are
    descriptive because retained pool sizes differ across procedures.
    $^\dagger$Prompt-matched IS rows are attribution controls.
    }
    \label{tab:app-vote-margins}
\end{table*}

Vote margins differ substantially by procedure. GPT-4o-mini IS
failures are rarely close contests: only one of its 18 strict vote
losses places the best correct workflow within two votes of the
selected wrong workflow. By contrast, 15 of 21 GPT-4o-mini MO-Tree
losses and 14 of 19 Qwen Run~1 MO-Tree losses fall within two votes.
Because retained pool sizes differ substantially across procedures,
these absolute margins are descriptive rather than directly comparable
effect sizes.

Across the nine primary pools, the median absolute vote deficit is
4 votes (mean 6.84). Including both prompt-matched controls changes
these values to 5 and 7.92 across all eleven pools.

\subsection{Vote-Loss Error Audit}
\label{app:error-audit}

\begin{table*}[t]
    \centering
    \small
    \renewcommand{\arraystretch}{1.15}
    \setlength{\tabcolsep}{4pt}
    \begin{tabular*}{\textwidth}{
        @{\extracolsep{\fill}}
        >{\raggedright\arraybackslash}p{0.11\textwidth}
        >{\raggedright\arraybackslash}p{0.18\textwidth}
        >{\raggedright\arraybackslash}p{0.58\textwidth}
        r@{}}
        \toprule
        \textbf{Type} & \textbf{Case} &
        \textbf{Selected divergence} & \textbf{Votes} \\
        \midrule
        Tool &
        GPT-4o-mini IS, Task~123 &
        \texttt{Building\_damage\_assessment}
        replaces \texttt{Change\_Mapping\_and\_Detection};
        remaining steps match. &
        22--2 \\
        Dependency &
        GPT-4o-mini IS, Task~179 &
        Three independent image restorations are changed into a
        sequential restoration chain. &
        20--11 \\
        Sem.\ equiv. &
        Qwen3.5-9B IS, Task~41 &
        Two independent steps are reordered while preserving tools,
        arguments, and dependency edges. &
        13--6 \\
        \bottomrule
    \end{tabular*}
    \caption{
    Representative cases from the 60-case semantic audit.
    Votes show selected workflow versus best exact-match workflow.
    The first two are substantive planning errors; the third is a
    semantics-preserving ordering difference rejected by strict Exact Match.
    }
    \label{tab:representative-errors}
\end{table*}

To determine whether strict Exact Match exaggerates the
discovery--selection gap, we audit non-tie vote-loss cases in which an
exact-match workflow survives structural filtering but a different
workflow receives strictly greater vote support.

The audit uses equal-allocation stratified sampling over six
model--procedure strata: GPT-4o-mini and Qwen3.5-9B Run~1 crossed with
IS, CoT-tree, and MO-Tree. We sample ten non-tie vote-loss cases from
each stratum, for 60 cases total.

Each selected workflow is classified according to its substantive
difference from the reference. The audit taxonomy contains:
\emph{Tool}, for an incorrect tool choice;
\emph{Dependency}, for incorrect information flow; and
\emph{Semantic Equivalent}, for a workflow that fails strict EM while
preserving the reference semantics. Representative examples of the three categories are shown in
Table~\ref{tab:representative-errors}.

\begin{table}[t]
    \centering
    \small
    \setlength{\tabcolsep}{4pt}
    \begin{tabular}{lrrr}
        \toprule
        Error Type & GPT-4o-mini & Qwen & Total \\
        \midrule
        Tool                 & 17 & 18 & 35 \\
        Dependency           & 13 &  9 & 22 \\
        Semantic Equivalent  &  0 &  3 &  3 \\
        \midrule
        Total                & 30 & 30 & 60 \\
        \bottomrule
    \end{tabular}

    \caption{
    Semantic audit of 60 non-tie vote-loss cases sampled from the
    primary pools. Fifty-seven contain substantive planning errors;
    the remaining three are semantics-preserving permutations of
    independent workflow steps.
    }
    \label{tab:app-error-audit}
\end{table}

Overall, 57 of 60 cases contain substantive planning errors
(95.0\%; Wilson 95\% CI: 86.3--98.3\%). All substantive failures fall
into two categories: incorrect tool selection (35 cases) or incorrect
dependency construction (22 cases).

The remaining three cases are semantically equivalent to the reference.
All occur in Qwen outputs and differ only by permutations of independent
workflow steps while preserving tools, arguments, and dependency
structure. An automatic permutation-equivalence detector identifies all
three and flags none of the other 57 cases. The detector is not intended
as a general workflow-graph isomorphism test.

Thus, strict-EM artifacts account for only a small fraction of the
sampled vote-loss failures. In most audited cases, the selected workflow
is substantively different from the reference even though a correct
workflow remains available in the structurally valid pool.

\subsubsection{Partial-Credit Characterization of Vote-Loss Winners}
\label{app:vote-loss-partial}

To complement the sampled semantic audit, we evaluate the selected
wrong workflow in every strict vote-loss case using DisasterBench's
tool-, parameter-, and dependency-level correctness checks. This
analysis covers all 265 strict vote-loss pool--task cases across the
eleven evaluated pools.

These metrics are task-level binary checks. Tool correctness requires
the predicted tool sequence to match the reference. Parameter
correctness requires exact agreement of every aligned step's inputs,
including input names, literal values, and references to upstream
generated outputs. Dependency correctness is likewise evaluated under
step-wise tool alignment. Exact-match workflows pass all three checks
by construction, so the zero parameter-correctness rate below reflects
the selected wrong workflows rather than an unattainable metric.

\begin{table}[t]
    \centering
    \small
    \setlength{\tabcolsep}{2pt}
    \begin{tabular*}{\columnwidth}{@{\extracolsep{\fill}}p{0.30\columnwidth}rrrr@{}}
        \toprule
        \shortstack{Correct-workflow\\rank} &
        $n$ & Tool & Parameter & Dependency \\
        \midrule
        Rank 2    & 151 & 29.8 & 0.0 & 2.6 \\
        Rank 3    & 55  & 25.5 & 0.0 & 1.8 \\
        Rank $>3$ & 59  & 23.7 & 0.0 & 1.7 \\
        \midrule
        All       & 265 & 27.5 & 0.0 & 2.3 \\
        \bottomrule
    \end{tabular*}

    \caption{
    Partial-credit correctness of the selected wrong workflow in strict
    vote-loss cases, grouped by the rank of the highest-ranked correct
    workflow. Entries are percentages of cases passing DisasterBench's
    task-level Tool, Parameter, and Dependency correctness checks.
    Parameter and Dependency correctness require step-wise tool
    alignment.
    }
    \label{tab:app-vote-loss-partial-rank}
\end{table}

Across all 265 cases, only 27.5\% of selected wrong workflows match the
reference tool sequence, and none achieve full parameter correctness.
Among the 73 wrong winners with the correct tool sequence, 68 fail the
parameter check because at least one generated-output reference is
misplaced or differs from the reference; the remaining five fail
because of literal input-value differences. Of these 68
reference-related cases, 67 also fail the step-level dependency check.
The remaining case preserves the correct upstream dependency but places
the generated output in the wrong input slot, so it fails parameter
correctness while passing dependency correctness. Together with the five literal-only cases, this yields six
dependency-correct but parameter-incorrect winners, corresponding to the 2.3\% reported in
Table~\ref{tab:app-vote-loss-partial-rank}.

The partial-credit profile changes little with the rank of the best
correct workflow: Tool correctness decreases only modestly from 29.8\%
at Rank~2 to 23.7\% below Rank~3. Thus, having a correct workflow near
the top of the vote ranking does not imply that the selected wrong
winner is itself a near-correct variant of the reference.

Viewed as mutually exclusive error patterns, 72.5\% of vote-loss
winners already have an incorrect tool sequence, 25.3\% preserve the
tool sequence but fail both parameter and dependency checks, 2.3\%
preserve both the tool sequence and dependencies but fail parameters,
and none differ only in dependencies.

\begin{table}[t]
    \centering
    \small
    \setlength{\tabcolsep}{4pt}
    \begin{tabular}{lrrrr}
        \toprule
        Vote-margin group & $n$ & Tool & Parameter & Dependency \\
        \midrule
        Low margin  & 133 & 18.8 & 0.0 & 3.0 \\
        High margin & 132 & 36.4 & 0.0 & 1.5 \\
        \bottomrule
    \end{tabular}

    \caption{
    Partial-credit correctness of selected wrong workflows split at the
    median normalized vote margin ($m=0.290$). High-margin wrong winners
    are more likely to preserve the reference tool sequence despite
    remaining incorrect in downstream parameter binding or dependency
    construction.
    }
    \label{tab:app-vote-loss-partial-margin}
\end{table}

The margin split reveals a counterintuitive pattern. High-margin wrong
winners are more likely to preserve the reference tool sequence than
low-margin wrong winners (36.4\% versus 18.8\%), while full parameter
correctness remains zero and dependency correctness remains rare. 
This suggests that strong vote support can persist even when
the selected workflow preserves the correct tool sequence but
contains incorrect parameter bindings or dependencies.

As a softer descriptive check not included in DisasterBench's official
metrics, 53.6\% of vote-loss winners have the same number of steps as
the reference, 61\% of tool positions match on average, and 14.0\%
have the same length while differing in exactly one tool position.
These statistics confirm that some wrong winners are locally close to
the reference, but most are not exact near-misses.

\subsubsection{Annotation Procedure}
\label{app:audit-annotation}

Following prior human--LLM collaborative evaluation practice
\citep{chen2026disastqa}, initial labels are generated by
\texttt{openai/gpt-6-astra-pro} with reasoning effort set to
\texttt{high}, using a fixed annotation rubric. For each case,
the model receives the task, reference workflow, selected winning
workflow, and relevant tool specifications, and assigns one of
the three error categories defined above.

Each of the three human reviewers separately examines all 60
LLM-generated labels against the task, reference workflow,
selected workflow, and relevant tool specifications.
All three confirm the error-category and semantic-equivalence
labels without revision.

Because the reviewers evaluate LLM-generated labels rather than
performing blind annotation from scratch, this result represents
unanimous human confirmation rather than an independent
inter-annotator agreement study. We therefore report the agreement
descriptively.

The audit annotation process and deployment verifier serve distinct roles.
The audit annotation process has access to the reference workflow for
post-hoc error classification, whereas the deployment verifier selects
among candidate workflows without access to the reference. Their results
are therefore not directly comparable as measures of verification
performance.

As a separate representation-level check, an automatic
permutation-equivalence detector identifies all three cases
labeled \emph{Semantic Equivalent} and none of the 57 cases
classified as substantive Tool or Dependency errors.

Cases are sampled equally across six model--procedure strata,
rather than proportionally to their frequency among all vote
losses. Accordingly, the Wilson confidence interval describes
uncertainty for the audited sample proportion and should not
be interpreted as a population-weighted estimate.

The annotation prompt and machine-readable audit records will
be released with the supplementary material.

\paragraph{Reference-consistency check.}

Two audited Tool-error cases introduce
\texttt{precipitation\_data\_convert\_tool}, but the additional step is
inconsistent with the task or declared tool interface. In one case, no
downstream \texttt{Flood\_depth\_prediction} step is requested and the
conversion output is unused. In the other, the converter---whose
declared input is radar echo frames---is applied to a rainfall CSV,
whereas comparable benchmark workflows pass that input directly to
\texttt{Flood\_depth\_prediction}.

An earlier fixed-seed case outside the primary analysis, GPT-4o-mini
MO-Tree Task~171, provides a useful contrast. Its workflow inserts the
same conversion tool between \texttt{Precipitation\_Nowcasting} and
\texttt{Flood\_depth\_prediction}. All other benchmark references with
this same nowcasting-to-flood-depth chain include the conversion step,
while Task~171 does not. We therefore treat Task~171 as a possible
reference inconsistency rather than as a substantive model error.

The distinction depends on consistency with the benchmark's task,
tool-interface, and workflow conventions rather than on the mere
presence of an additional step.

\subsection{Concentration and Recurrence of Winning Errors}
\label{app:modal-errors}

The semantic audit establishes that most selected wrong workflows are
substantively incorrect. We next characterize two distinct properties:
\emph{concentration}, which measures how strongly the selected wrong
workflow dominates the valid vote distribution, and
\emph{recurrence}, which measures whether the same error reappears
across procedures or independent runs.

\paragraph{Vote concentration.}

Among GPT-4o-mini vote-loss cases, the mean winning wrong-workflow
share is .68 for IS, .41 for CoT-tree, and .41 for MO-Tree. The
strongest correct workflow receives mean shares of .15, .15, and .22,
respectively. The corresponding mean absolute winning margins are
16.3, 8.1, and 2.2 votes.

For Qwen Run~1, the wrong-winner shares are .55, .43, and .54 for IS,
CoT-tree, and MO-Tree, with strongest-correct shares of .17, .14, and
.23. For Run~2, the corresponding wrong-winner shares are .54, .45,
and .57, with strongest-correct shares of .18, .17, and .21.

Absolute margins and winner shares are not directly comparable across
procedures because retained pool sizes differ substantially. We
therefore use these quantities descriptively rather than as
cross-procedure causal measures.

\paragraph{Cross-procedure recurrence.}

We first restrict attention to tasks on which both procedures incur a
strict vote loss and compare their exact normalized wrong winner.

\begin{table}[t]
    \centering
    \small
    \setlength{\tabcolsep}{3.5pt}
    \begin{tabular}{lrrr}
        \toprule
        Model / Run &
        IS--CoT &
        IS--MO &
        CoT--MO \\
        \midrule
        GPT-4o-mini & 3/4   & 4/6 & 3/6 \\
        Qwen Run 1  & 14/15 & 5/5 & 6/7 \\
        Qwen Run 2  & 13/15 & 4/6 & 3/5 \\
        \bottomrule
    \end{tabular}

    \caption{
    Exact agreement of the selected incorrect workflow on tasks where
    both procedures incur a strict vote loss. Entries show same wrong
    winner / shared vote-loss tasks.
    }
    \label{tab:app-modal-agreement}
\end{table}

The GPT-4o-mini shared-failure sets contain only four to six tasks, so
their agreement fractions have wide uncertainty and should not be used
to define a model-level recurrence regime.

Qwen provides stronger evidence. In Run~1, IS and CoT-tree select the
same wrong workflow on 14 of 15 shared vote-loss tasks; IS and
MO-Tree agree on all five; and CoT-tree and MO-Tree agree on six of
seven. Run~2 again shows strong IS--CoT agreement at 13 of 15,
whereas the smaller pairs involving MO-Tree agree on 4 of 6 and 3 of
5 tasks.

A broader asymmetric comparison yields the same contrast. When
CoT-tree incurs a vote loss, IS selects the same incorrect workflow
on 5/22 GPT-4o-mini cases (23\%), compared with 20/28 Qwen Run~1
cases (71\%) and 17/22 Run~2 cases (77\%). Conversely, when IS incurs
a vote loss, CoT-tree selects the same wrong winner on 5/18
GPT-4o-mini cases (28\%), 18/28 Qwen Run~1 cases (64\%), and 22/33
Run~2 cases (67\%).

\paragraph{Cross-run recurrence.}

The two independent Qwen runs permit a cleaner stability analysis
within the same generation procedure. IS incurs a strict vote loss on
the same 25 tasks in both runs and selects the identical wrong workflow
on 24 of them (96\%). CoT-tree shares 19 vote-loss tasks across runs
and repeats the same wrong winner on 15 (79\%). MO-Tree shares six
vote-loss tasks and repeats the same wrong winner on three (50\%).

Notably, the IS--CoT cross-procedure agreement in Qwen Run~1
(14/15, 93\%) is only slightly lower than IS's own across-run
recurrence (24/25, 96\%). Because the realized CoT-tree configuration
is non-adaptive, this comparison primarily reflects a change in
generation scaffold rather than the introduction of adaptive tree
search.

Whether winning errors recur appears to track how strongly the
generation scaffold changes the candidate distribution. Qwen IS and
CoT-tree exhibit highly similar modal errors, whereas their
GPT-4o-mini error distributions overlap substantially less.

\subsection{Selector Sensitivity}
\label{app:selector-sensitivity}

The primary selector preserves the multiplicity semantics of each
generation procedure. For IS, repeated retained generations
contribute repeated vote mass; CoT-tree and MO-Tree instead use their
implemented retention or aggregation rules to construct the pool
presented to selection.

We test whether the main results depend on this convention using
several deterministic alternatives:
\emph{unfiltered} selection, which omits structural filtering;
\emph{occurrence-weighted} voting;
\emph{unique-text} voting, which removes repeated identical raw
candidate texts before voting;
\emph{first-valid} selection; and an
\emph{oracle tie-break} diagnostic, which assumes that every
highest-vote tie containing a correct workflow is resolved correctly.

Unique-text voting removes duplicates before normalization and workflow
grouping. Thus, distinct generated texts that normalize to the same
workflow still contribute separate votes. Occurrence-weighted voting
assigns one vote per generation call for IS and CoT-tree and one vote
per retained solution node for MO-Tree; multiplicity already aggregated
within an MO-Tree solution node cannot be reconstructed.

\begin{table*}[t]
    \centering
    \small
    \setlength{\tabcolsep}{3.2pt}
    \begin{tabular}{llrrrrrr}
        \toprule
        Model / Run & Procedure &
        Primary &
        Unfiltered &
        Occurrence &
        Unique text &
        First valid &
        Tie upper bound \\
        \midrule

        GPT-4o-mini
        & IS      & 121 & 121 & 121 & 119 & 119 & 121 \\
        & CoT-tree & 131 & 129 & 130 & 131 & 112 & 131 \\
        & MO-Tree  & 123 & 122 & 123 & 123 & 104 & 125 \\

        \midrule

        Qwen Run 1
        & IS      & 129 & 129 & 129 & 128 & 121 & 131 \\
        & CoT-tree & 132 & 132 & 127 & 132 & 122 & 134 \\
        & MO-Tree  & 122 & 122 & 127 & 122 & 114 & 128 \\

        \midrule

        Qwen Run 2
        & IS      & 128 & 128 & 128 & 130 & 119 & 129 \\
        & CoT-tree & 133 & 133 & 128 & 133 & 118 & 135 \\
        & MO-Tree  & 122 & 121 & 122 & 122 & 110 & 129 \\

        \bottomrule
    \end{tabular}

    \caption{
    Deterministic selector sensitivity on the nine primary pools.
    Entries are numbers of correctly selected tasks out of 183.
    Unique-text voting removes duplicate raw generations before
    normalization. The tie upper bound assumes that every highest-vote
    tie containing a correct workflow is resolved in its favor.
    }
    \label{tab:app-voting-sensitivity}
\end{table*}

The deterministic variants are highly stable. Removing structural
filtering or switching between occurrence-weighted and unique-text
voting changes accuracy by only 0--5 tasks across the primary pools,
with no statistically significant differences. Even an oracle
tie-break recovers only 0--7 additional tasks.

The clear exception is first-valid selection. For GPT-4o-mini,
CoT-tree and MO-Tree each lose 19 correct tasks relative to their
primary selectors ($p<.001$). Candidate-frequency information
therefore carries useful information. The selection problem is not
that majority voting is intrinsically poor, but that even a competitive
majority-based selector leaves substantial Oracle coverage unrealized.

\paragraph{GPT-4o judge.}

We additionally evaluate \texttt{openai/gpt-4o} as a generic
model-based selector. The judge receives the structurally valid pool
after deduplication by normalized workflow. If more than eight distinct
valid workflows remain, eight are sampled uniformly using a fixed
task-specific seed; candidate labels are then shuffled and blinded.
Tasks containing only one distinct valid workflow require no judge
call.

The judge uses temperature 0 and a maximum output length of 300 tokens.
For the three GPT-4o-mini candidate pools, all 360 judge calls terminate
normally, with at most 61 completion tokens, and every response parses
to a valid candidate label; the majority fallback is never invoked.
The same holds for the six Qwen pools: all 745 judge decisions parse
successfully, no fallback is used, and the corresponding calls terminate
normally with at most 71 completion tokens.

More than eight distinct valid workflows remain on 7, 27, and 23 tasks
in the GPT-4o-mini IS, CoT-tree, and MO-Tree pools, respectively, so
these cases are evaluated on a uniformly sampled subset of eight
workflows.

Because the judge observes at most eight deduplicated workflows and
does not receive the candidate multiplicity available to the primary
majority selector, we treat it as a selector-sensitivity diagnostic
rather than an exhaustive or compute-matched verifier.

\begin{table*}[t]
    \centering
    \small
    \setlength{\tabcolsep}{4pt}
    \begin{tabular}{llrrrr}
        \toprule
        Model / Run & Procedure &
        Primary &
        GPT-4o judge &
        Tie upper bound &
        Oracle \\
        \midrule

        GPT-4o-mini
        & IS      & 121 & 123 & 121 & 139 \\
        & CoT-tree & 131 & 112 & 131 & 153 \\
        & MO-Tree  & 123 & 112 & 125 & 146 \\

        \midrule

        Qwen Run 1
        & IS      & 129 & 110 & 131 & 159 \\
        & CoT-tree & 132 & 114 & 134 & 162 \\
        & MO-Tree  & 122 & 128 & 128 & 147 \\

        \midrule

        Qwen Run 2
        & IS      & 128 & 106 & 129 & 162 \\
        & CoT-tree & 133 & 115 & 135 & 157 \\
        & MO-Tree  & 122 & 121 & 129 & 147 \\

        \bottomrule
    \end{tabular}

    \caption{
    GPT-4o judge and oracle tie-break diagnostics on the nine primary
    candidate pools. The judge observes at most eight deduplicated
    workflows per task, whereas the primary selector uses candidate
    multiplicity in the realized pool.
    }
    \label{tab:app-judge}
\end{table*}

The judge does not provide a consistent improvement over majority
selection. It improves GPT-4o-mini IS from 121 to 123 correct tasks
and Qwen Run~1 MO-Tree from 122 to 128, but degrades several other
pools substantially. In particular, GPT-4o-mini CoT-tree falls from
131 to 112 tasks ($p=.001$), while the larger Qwen IS and CoT-tree
pools lose 18--22 correct tasks relative to the primary selector.

One plausible explanation is information loss at the judge interface:
candidate multiplicity is removed, and tasks with more than eight
distinct workflows are further reduced to a sampled subset. We
therefore interpret these results as sensitivity to this particular
judge interface rather than as evidence against model-based
verification more generally.

Selector choice can also reverse apparent procedure rankings. Under the
primary selector, Qwen MO-Tree trails IS by seven and six tasks in the
two runs. Under the GPT-4o judge, MO-Tree instead exceeds IS by 18
tasks ($p=.002$) and 15 tasks ($p=.028$). For GPT-4o-mini, CoT-tree
exceeds IS by 10 tasks under the primary selector but trails IS by
11 tasks under the judge. Judge-based rankings should therefore be
interpreted strictly as selector sensitivity rather than evidence
favoring a candidate-generation procedure.

\subsection{Targeted Top-3 Semantic Verification}
\label{app:targeted-verifier}

\paragraph{Oracle-conditioned diagnostic.}

We first evaluate the verifier on the 181 primary-pool cases in which
majority selection is incorrect and an exact-match workflow appears
among the first three ranked workflow groups. Candidate labels are
shuffled and blinded, and the verifier receives the task, the Top-3
workflows, and the full specifications of all relevant tools, but not
vote counts. It is instructed to choose the workflow that best
satisfies the task while respecting the declared tool interfaces and
cross-step information flow; it is not given the reference workflow or
any Exact-Match signal. The verifier is GPT-4o
(\texttt{openai/gpt-4o} via OpenRouter), the same model used for the
generic judge, with temperature 0, no request seed, and a maximum
output length of 300 tokens. 
Thirty-four cases contain only two distinct ranked groups.
Verifier calls are a separate post-generation selection cost, excluded
from candidate-generation budgets. The 181 diagnostic calls consume
321,686 tokens (1,777 per call on average). Deployable evaluation
reuses these decisions and makes 924 additional calls, consuming
1,574,254 tokens. Across both stages, all 1,105 calls succeed without
retries or truncation, consuming 1,895,940 tokens in total.

The verifier selects the correct workflow in 90 of 181 cases (49.7\%),
compared with 66.0 cases (36.5\%) expected under uniform random choice
over each case's available candidates. Recovery is higher when the
correct workflow ranks second (76/137; 55.5\%) than when it ranks third
(14/44; 31.8\%). Richer semantic information therefore improves
discrimination in known-recoverable cases but still leaves about half
unresolved.

\paragraph{Deployable protocol.}

We next apply the same interface to all 183 tasks in each of the nine
primary pools. Tasks with one valid workflow group are left unchanged;
for each task with at least two groups, the verifier selects among the
first three ranked groups. This yields 1,105 contested pool--task
cases.

Relative to majority selection, the verifier fixes 90 errors but
breaks 175 already-correct decisions, for a net change of $-85$ across
the nine primary pools. These 90 fixes are exactly the 90 recoveries
identified in the oracle-conditioned diagnostic because the deployable
selector reuses the cached verifier decisions on identical inputs.
Eight of nine pools decline. The main failure mode is therefore false
override: the verifier can recognize some majority errors but too often
replaces already-correct high-support decisions.

\begin{table*}[t]
    \centering
    \small
    \setlength{\tabcolsep}{3pt}
    \begin{tabular}{llrrrrrr}
        \toprule
        Model / Run & Procedure &
        Majority & Verifier & Fixed & Broken & Net & McNemar $p$ \\
        \midrule

        GPT-4o-mini & IS      & 121 & 123 & 11 & 9  & +2  & .820 \\
        GPT-4o-mini & CoT-tree & 131 & 113 & 3  & 21 & -18 & $<.001$ \\
        GPT-4o-mini & MO-Tree  & 123 & 108 & 12 & 27 & -15 & .024 \\

        \midrule

        Qwen Run 1 & IS      & 129 & 116 & 9  & 22 & -13 & .029 \\
        Qwen Run 1 & CoT-tree & 132 & 117 & 11 & 26 & -15 & .020 \\
        Qwen Run 1 & MO-Tree  & 122 & 121 & 12 & 13 & -1  & 1.000 \\

        \midrule

        Qwen Run 2 & IS      & 128 & 118 & 13 & 23 & -10 & .130 \\
        Qwen Run 2 & CoT-tree & 133 & 122 & 8  & 19 & -11 & .052 \\
        Qwen Run 2 & MO-Tree  & 122 & 118 & 11 & 15 & -4  & .560 \\

        \bottomrule
    \end{tabular}

    \caption{
    Deployable targeted Top-3 verifier results on the nine primary
    pools. Fixed counts majority-incorrect cases corrected by the
    verifier; Broken counts majority-correct cases overturned
    incorrectly. McNemar tests are computed separately within each
    pool.
    }
    \label{tab:app-targeted-verifier}
\end{table*}

\subsubsection{Why Does the Verifier Produce False Overrides?}

We compare corrected majority errors with incorrect overrides among
changed decisions in the deployable protocol.

\begin{table}[t]
    \centering
    \small
    \renewcommand{\arraystretch}{1.08}
    \setlength{\tabcolsep}{3pt}
    \begin{tabular*}{\columnwidth}{@{\extracolsep{\fill}}p{0.52\columnwidth}rr@{}}
        \toprule
        Statistic & Fixes (90) & \shortstack{False overrides\\(175)} \\
        \midrule
        Median normalized margin & 0.258 & 0.500 \\
        Median winner vote share & 0.556 & 0.700 \\
        Median valid groups & 3 & 4 \\
        Chosen rank 2 & 84.4\% & 64.6\% \\
        Chosen rank 3 & 15.6\% & 35.4\% \\
        More steps than majority & 27.8\% & 52.6\% \\
        \bottomrule
    \end{tabular*}
    \caption{
    Characteristics of changed decisions under deployable targeted
    Top-3 verification, comparing corrected majority errors with false
    overrides.
    }
    \label{tab:app-verifier-changed-decisions}
\end{table}

\begin{table}[t]
    \centering
    \small
    \renewcommand{\arraystretch}{1.08}
    \setlength{\tabcolsep}{3pt}
    \begin{tabular}{lrrrrr}
        \toprule
        \shortstack{Margin\\bin} &
        \shortstack{Majority\\correct} &
        Broken &
        \shortstack{Break rate\\(\%)} &
        Fixed & Net \\
        \midrule
        $[0,0.1)$   & 51  & 22 & 43.1 & 21 & -1 \\
        $[0.1,0.2)$ & 39  & 9  & 23.1 & 16 & +7 \\
        $[0.2,0.3)$ & 60  & 18 & 30.0 & 15 & -3 \\
        $[0.3,0.5)$ & 94  & 28 & 29.8 & 20 & -8 \\
        $[0.5,1.0]$ & 394 & 98 & 24.9 & 18 & -80 \\
        \bottomrule
    \end{tabular}
    \caption{
    Deployable verifier outcomes by normalized vote-margin bin across
    the nine primary pools. Break rate is Broken divided by the number
    of majority-correct cases in each bin; Net is Fixed minus Broken.
    }
    \label{tab:app-verifier-margin-bins}
\end{table}

False overrides are not confined to small vote margins. High-margin
cases contain many more already-correct majority decisions than
recoverable errors, so extending verifier intervention into this region
produces a strongly asymmetric cost. Among the 175 false overrides,
the replacement workflow passes the Tool, Parameter, and Dependency
checks in only 20.6\%, 2.9\%, and 4.0\% of cases, respectively.
Moreover, 52.6\% of false overrides add steps relative to the correct
majority, compared with 27.8\% among fixes.

\paragraph{Margin-gated verifier intervention.}

Because the ungated verifier is not given candidate-support
information, we test whether it can be used more selectively without
additional model calls. Let $v_1$ and $v_2$ denote the vote counts of
the first- and second-ranked workflow groups and $V$ the total number
of valid retained votes. We define

\[
m=\frac{v_1-v_2}{V}.
\]

For threshold $t$, the verifier may override majority selection only
when $m\leq t$; otherwise the majority winner is retained. All verifier
decisions are reused from the cached 1,105 contested cases.

\begin{table*}[t]
    \centering
    \small
    \setlength{\tabcolsep}{4pt}
    \begin{tabular}{rrrrrrrr}
        \toprule
        Threshold $t$ &
        Routed &
        Changed &
        Fixed &
        Broken &
        Net &
        Selected &
        McNemar $p$ \\
        \midrule

        0        & 92   & 49  & 10 & 9   & +1  & 1142 & 1.000 \\
        .05      & 114  & 65  & 13 & 13  & 0   & 1141 & 1.000 \\
        .10      & 190  & 118 & 22 & 24  & -2  & 1139 & .880 \\
        .15      & 243  & 147 & 31 & 27  & +4  & 1145 & .690 \\
        .20      & 321  & 178 & 37 & 34  & +3  & 1144 & .810 \\
        .30      & 437  & 231 & 53 & 51  & +2  & 1143 & .920 \\
        .40      & 542  & 271 & 66 & 63  & +3  & 1144 & .860 \\
        .50      & 652  & 321 & 76 & 88  & -12 & 1129 & .390 \\
        $\infty$ & 1105 & 455 & 90 & 175 & -85 & 1056 & $<.001$ \\

        \bottomrule
    \end{tabular}

    \caption{
    Margin-gated deployment of the targeted Top-3 verifier across the
    nine primary pools. The majority baseline selects 1,141
    pool--task cases correctly. Routed counts cases satisfying
    $m\leq t$; Changed counts cases in which the verifier differs from
    majority selection. Thresholds are evaluated post hoc and are
    therefore sensitivity analyses rather than tuned deployment
    results.
    }
    \label{tab:app-margin-gated-verifier}
\end{table*}

Restricting verifier intervention to low-margin cases removes most of
the degradation of unrestricted deployment. The best post-hoc
threshold, $t=.15$, yields a descriptive gain of four pool--task cases
over majority selection (1,145 versus 1,141), but the difference is not
significant ($p=.690$) and the threshold is selected on the evaluation
set.

Leave-one-run-out threshold selection removes this apparent gain.
Across the three held-out folds, the gated selector fixes 38 errors and
breaks 39 correct majority decisions, for a net change of $-1$
pool--task case. An analogous sweep using the absolute vote margin
$\Delta=v_1-v_2$ yields the same qualitative result, with an aggregate
leave-one-run-out change of $-2$.

Candidate-support information therefore identifies regions in which
unrestricted semantic override is dangerous, but low vote margin does
not identify a subset on which the current verifier reliably
outperforms majority selection. The remaining Oracle headroom cannot
be recovered by simply combining this verifier with a fixed
vote-margin gate.

The verifier prompt and per-case decision records are included in the
supplementary material.

\subsection{Representative Workflow Case}
\label{app:task22-case}

\paragraph{Dependency error: Task~22.}

Task~22 provides a concrete example of how a change in generation
scaffold can reshape a dominant dependency error. This case is analyzed
separately from the stratified 60-case semantic audit.

The task requires generating a high-resolution image and applying both
geospatial object segmentation and landslide segmentation. Let
$X_{\mathrm{img}}$ denote the generated high-resolution image,
$T_{\mathrm{obj}}$ the object-segmentation step, and
$T_{\mathrm{land}}$ the landslide-segmentation step. The reference
dependency structure is

\[
X_{\mathrm{img}} \rightarrow T_{\mathrm{obj}},
\qquad
X_{\mathrm{img}} \rightarrow T_{\mathrm{land}}.
\]

A common incorrect workflow instead introduces the sequential
dependency

\[
X_{\mathrm{img}}
\rightarrow T_{\mathrm{obj}}
\rightarrow T_{\mathrm{land}}.
\]

\begin{figure*}[t]
    \centering
    \includegraphics[width=\textwidth]{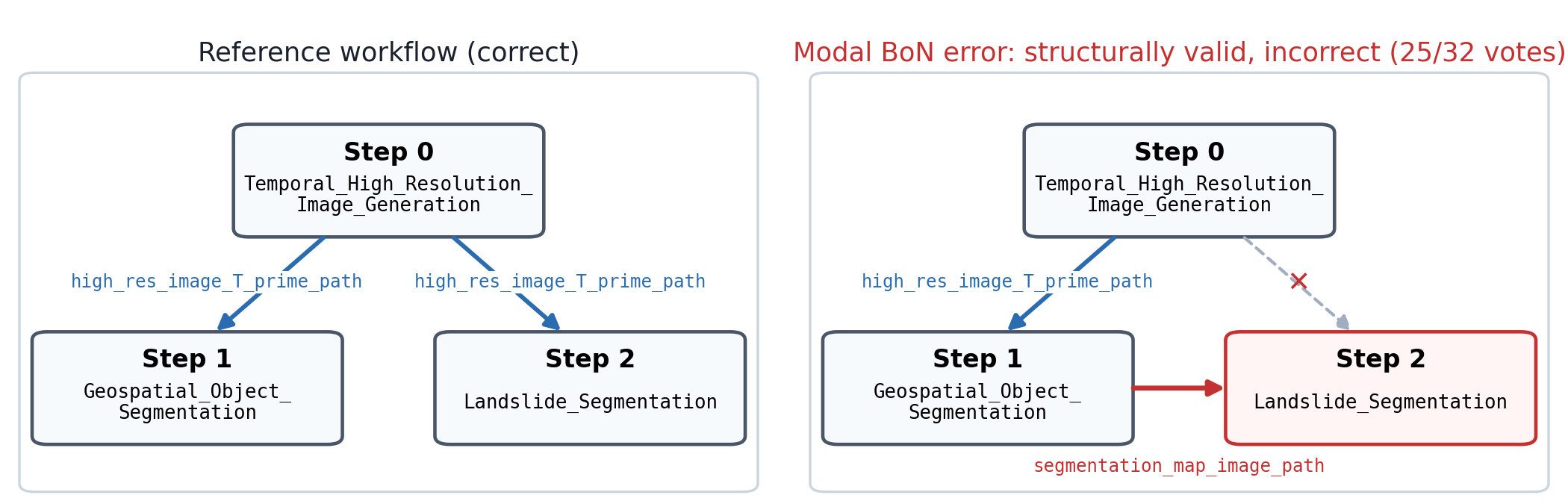}
    \caption{
    Representation-level validity versus planning correctness on
    Task~22. The reference sends the generated image independently to
    both downstream tools, whereas the dominant IS error routes the
    object-segmentation output into the landslide-segmentation step.
    The latter is structurally well formed but changes the intended
    information flow.
    }
    \label{fig:workflow-representation-example}
\end{figure*}

The incorrect workflow is structurally well formed and passes the
primary validator, but it changes the intended information flow.
\texttt{Landslide\_Segmentation} expects a high-resolution aerial or
satellite image, including a reconstructed image from a previous step.
The incorrect workflow instead supplies the segmentation map produced
by \texttt{Geospatial\_Object\_Segmentation}.

\begin{table}[t]
  \centering
  \scriptsize
  \setlength{\tabcolsep}{3pt}
  \begin{tabular*}{\columnwidth}{@{\extracolsep{\fill}}lrrl@{}}
    \toprule
    Pool & \shortstack{Correct\\votes} &
    \shortstack{Wrong-dep.\\votes} & Result \\
    \midrule
    IS                & 7  & 25 & Wrong \\
    Prompt-matched IS & 29 & 1  & Correct \\
    CoT-tree           & 28 & 2  & Correct \\
    MO-Tree            & 4  & 3  & Correct \\
    \bottomrule
  \end{tabular*}

  \caption{
  Candidate support for the correct workflow and dominant
  dependency-error alternative on GPT-4o-mini Task~22. Remaining
  votes, when present, correspond to other workflow variants.
  }
  \label{tab:app-task22}
\end{table}

In IS, the dependency-error workflow receives 25 of 32 valid votes,
compared with seven for the correct workflow. Prompt-matched IS nearly
reverses this distribution without tree search: 29 votes support the
correct workflow and only one reproduces the dependency error.
CoT-tree behaves similarly, with 28 correct votes and two votes for the
same error. MO-Tree also selects the correct workflow.

Task~22 therefore provides a concrete example in which changing the
generation scaffold, without introducing adaptive tree search, is
sufficient to suppress a dominant dependency-error mode. It also
illustrates the distinction between structural validity and planning
correctness: the IS winner is representation-level valid but routes
information through the wrong intermediate result.

\section{Full Attribution Checklist}
\label{app:full-checklist}

Table~\ref{tab:app-full-checklist} expands the compact checklist in
Table~\ref{tab:attribution-checklist} with diagnostic criteria,
empirical evidence, and implementation guidance.

\begin{table*}[t]
    \centering
    \scriptsize
    \renewcommand{\arraystretch}{1.15}
    \setlength{\tabcolsep}{3pt}
    \begin{tabular*}{\textwidth}{
        @{\extracolsep{\fill}}
        p{0.12\textwidth}
        p{0.20\textwidth}
        p{0.20\textwidth}
        p{0.25\textwidth}
        p{0.17\textwidth}
        @{}}
        \toprule
        \textbf{Factor} &
        \textbf{How it arises} &
        \textbf{How to detect it} &
        \textbf{Observed effect in this study} &
        \textbf{Alignment principle} \\
        \midrule

        Generation scaffold
        &
        Nominally different inference procedures use different prompts,
        reasoning instructions, or expansion templates.
        &
        Reproduce the generation scaffold with independent sampling while
        holding decoding, budget, and downstream evaluation fixed.
        &
        Prompt-matched IS reaches CoT-tree-level Oracle coverage on
        GPT-4o-mini (154 versus 153) and lies within the two Qwen
        CoT-tree runs (160 versus 162 and 157)
        (Appendix~\ref{app:prompt-matched}).
        &
        Match scaffolds or include a scaffold-matched sampling control.
        \\

        \midrule

        Sampling controls
        &
        Request seeds and other sampling settings can alter response
        diversity under repeated or identical prompts.
        &
        Use matched reruns where possible and distinguish controlled seed
        effects from broader sampling-sensitivity comparisons.
        &
        In a matched IS rerun, removing the fixed request seed increases
        distinct raw responses from 2.03 to 3.96 and unique valid
        workflows from 1.60 to 2.32 per task; Oracle changes from
        133 to 139 ($p=.11$)
        (Appendix~\ref{app:seed}).
        &
        Apply sampling controls symmetrically and distinguish matched
        controls from broader sensitivity checks.
        \\

        \midrule

        Post-generation processing
        &
        Candidate pools receive different repair, normalization, or
        semantic-correction rules.
        &
        Re-extract raw outputs and reevaluate all procedures through one
        common transformation pipeline.
        &
        On the primary GPT-4o-mini pools, the common recoverable
        normalization raises CoT-tree Oracle from 145 to 153 while
        leaving IS unchanged at 139. On earlier fixed-seed pools,
        procedure-specific processing yields 27.4\% CoT-tree invalidity
        versus 2.9\% under the common policy, and unfiltered-majority
        correctness rises from 104 to 118
        (Appendix~\ref{app:transformation-audit}).
        &
        Use one audited normalization policy and separate recoverable
        canonicalization from semantic correction.
        \\

        \midrule

        Backend realization
        &
        Serving systems may not realize requested multi-sample or
        branching parameters.
        &
        Audit returned completions and execution logs rather than relying
        on nominal hyperparameters.
        &
        GPT-4o-mini returns one completion for nominal $n>1$ requests,
        whereas Qwen realizes MO-Tree's configured local multi-sample
        aggregation
        (Appendix~\ref{app:backend}).
        &
        Report realized branching and avoid treating backend differences
        as model-only effects.
        \\

        \midrule

        Resource accounting
        &
        Retained candidates may represent only part of the inference used
        to construct a trajectory or pooled candidate set.
        &
        Sum all model calls and tokens contributing to the evaluated pool,
        including generations later discarded or aggregated.
        &
        An exploratory pooled configuration was initially estimated at
        approximately 232k tokens per task from its retained subset, but
        full accounting across its contributing runs yields approximately
        414k tokens per task, about $1.74\times$ the IS budget
        (Appendix~\ref{app:pooled-accounting}).
        &
        Account for full realized inference cost rather than retained-pool
        size.
        \\

        \midrule

        Selection interface
        &
        Different selectors observe different multiplicities, subsets, or
        representations of the candidate pool.
        &
        Reevaluate fixed pools with alternative deterministic selectors
        and explicitly specified verifier interfaces.
        &
        Deterministic selector variants are largely stable, but
        model-based selectors do not reliably improve on majority
        selection. The targeted Top-3 verifier fixes 90 majority errors
        while breaking 175 correct majority decisions under unrestricted
        deployment; vote-margin gating removes most of this degradation
        but yields no reliable improvement
        (Appendix~\ref{app:selector-sensitivity};
        Appendix~\ref{app:targeted-verifier}).
        &
        Hold the selector fixed for primary comparisons, preserve
        candidate-support information, and report selector sensitivity
        separately.
        \\

        \bottomrule
    \end{tabular*}

    \caption{
    Full attribution checklist for structured test-time inference.
    Each factor can alter the realized candidate pool or measured
    outcome without being part of the nominal algorithmic distinction
    under study.
    }
    \label{tab:app-full-checklist}
\end{table*}

\section{Extended Related Work and Positioning}
\label{app:related-work}

This appendix provides additional context for the literature discussed
in Section~\ref{sec:related-work} and clarifies how our study differs
from prior work on test-time search, verification, and structured-agent
evaluation.

\subsection{Test-Time Compute, Search, and Verification}

Test-time reasoning methods improve candidate generation through
repeated sampling, self-consistency, and structured search
\citep{wang2022self,yao2023tree,hao2023reasoning,wu2024beyond}.

\citet{wang2024reasoning} study reasoning strategies under explicit
inference budgets and show that the apparent advantage of more complex
methods can shrink or disappear when compared against stronger
compute-matched baselines.
Their analysis establishes inference cost as a critical comparison
variable.

Recent system-level treatments of test-time scaling further emphasize
that inference regimes should be compared with explicit attention to
compute accounting, evaluation protocol, and reproducibility
\citep{hariri2026testtime}.
Our analysis is complementary: rather than proposing a general taxonomy
of inference regimes, we use matched empirical controls to identify
which components of a realized structured-planning pipeline account for
an observed performance difference.

We control realized inference expenditure but additionally ask whether
remaining differences between nominal procedures are caused by the
search structure itself or by other components of the realized
pipeline.
In our GPT-4o-mini comparison, for example, the original IS and
CoT-tree configurations already consume similar inference budgets,
while a prompt-matched independent-sampling control reproduces the
CoT-tree discovery result.
This isolates generation scaffold as an additional attribution
variable beyond compute alone.

Prior work also emphasizes verification, selection, and budget control.
\citet{chen2024tree} analyze how discriminator accuracy affects tree
search, iterative correction, and re-ranking.
\citet{liu2026bag} benchmark proposer--verifier inference-time
computation while varying candidate-generation and reward mechanisms.
\citet{suh2026think} combine Best-of-$N$ sampling and beam search in a
hybrid inference pipeline, while \citet{fang2026inference} formulate
two-stage inference-time budget control over search actions and final
answer commitment.
Finally, \citet{wu2024beyond} use MCTS to explore reasoning paths
constructed from five atomic high-level reasoning actions.

Most directly, \citet{brown2024large} show that candidate coverage can
continue increasing under repeated sampling while common selection
methods, including majority voting and reward-model selection, plateau.
We therefore do not claim the general discovery--selection gap itself
as novel.
Our focus is instead on how this gap interacts with attribution in
typed workflow planning:
whether an apparent gain should be attributed to search structure,
generation scaffold, or downstream evaluation.

This distinction becomes important when nominally different methods
also differ in their prompts, retention rules, or serving behavior.
In our GPT-4o-mini experiments, for example, an apparent CoT-tree
Oracle advantage over IS is reproduced by independent sampling using
the same CoT-tree expansion prompt.
The relevant question is therefore not simply whether one nominal
inference procedure outperforms another, but which component of the
realized inference pipeline produces the observed difference.

\subsection{Evaluation Pipelines for Structured Agents}

\citet{zhou2026toolgrad} construct valid tool-use chains before
synthesizing corresponding user requests, illustrating the importance
of explicit structural constraints in tool-use data construction.
Our setting asks a different question: whether a structurally valid
generated workflow is also the correct planning decision for the task.

Recent agent-evaluation work emphasizes that framework configuration,
runtime realization, and evaluation interfaces can materially affect
measured system performance
\citep{yao2026harness,zhu2025oagents,liu2026benchmarking}.
Harness-Bench, for example, studies configuration-level harness effects
across realistic agent workflows under shared task environments and
evaluation protocols~\citep{yao2026harness}.

Our setting is complementary: rather than benchmarking executable agent
frameworks, we audit attribution within offline structured planning.
We explicitly align or audit generation scaffold, sampling behavior,
post-generation processing, backend realization, resource accounting,
and selection, and then separate candidate discovery from structural
validity and final selection.

\subsection{Disaster-Agent and Workflow Planning}

Our experiments build on DisasterBench
\citep{chen2026disasterbench}, which evaluates planning over
26 disaster-response tools with fixed input--output interfaces.
The benchmark contains 233 tasks, of which we use the 183-task
evaluation split.
Each task requires a structured workflow that selects tools, binds
task-specific parameters, and specifies dependencies between steps.

The original benchmark evaluates planning using Exact Match together
with tool-, parameter-, and dependency-level accuracy and compares
multiple planning paradigms including Direct, CoT, ToT, RAP, and
ReAct.
It further analyzes the first point of failure (FPoF) in incorrect
workflows.

The present study asks a different question from the original
DisasterBench evaluation.
DisasterBench diagnoses where an individual predicted workflow first
fails in tool selection, parameter binding, or dependency construction.
Here, we instead study multi-candidate test-time inference: whether a
correct workflow is present in the candidate pool, whether it survives
structural validation, and why it may still fail to be selected.
The benchmark's typed representation makes these stages and their
associated errors directly observable.

\subsection{Positioning Against Prior Work}

Table~\ref{tab:related-positioning} summarizes the dimensions most
relevant to our study.
A checkmark indicates that a dimension is explicitly studied.

\begin{table*}[t]
  \centering
  \scriptsize
  \renewcommand{\arraystretch}{1.12}
  \setlength{\tabcolsep}{3pt}
  \begin{tabular*}{\textwidth}
  {@{\extracolsep{\fill}}p{0.34\textwidth}cccccccc@{}}
    \toprule
    Work & (a) & (b) & (c) & (d) & (e) & (f) & (g) & (h) \\
    \midrule

    ToT / RAP / HiAR-style search
    \citep{yao2023tree,hao2023reasoning,wu2024beyond}
    &  & $\surd$ &  &  &  &  &  &  \\

    Budget-aware reasoning comparison
    \citep{wang2024reasoning}
    &  & $\surd$ & $\surd$ &  &  &  &  &  \\

    System-level test-time scaling
    \citep{hariri2026testtime}
    &  & $\surd$ & $\surd$ & $\surd$ & $\surd$ &  &  &  \\

    Discriminator-dependent tree search
    \citep{chen2024tree}
    &  & $\surd$ &  &  & $\surd$ &  &  &  \\

    Repeated-sampling coverage
    \citep{brown2024large}
    &  &  & $\surd$ & $\surd$ & $\surd$ &  &  &  \\

    Candidate generation and verification
    \citep{liu2026bag}
    &  & $\surd$ & $\surd$ &  & $\surd$ &  &  &  \\

    Hybrid sampling and search
    \citep{suh2026think}
    &  & $\surd$ &  &  &  &  &  &  \\

    Inference-time budget allocation
    \citep{fang2026inference}
    &  &  & $\surd$ &  &  &  &  &  \\

    Structured tool-use construction
    \citep{zhou2026toolgrad}
    & $\surd$ &  &  &  &  &  &  &  \\

    Agent evaluation frameworks
    \citep{yao2026harness,zhu2025oagents,liu2026benchmarking}
    &  &  &  &  &  & $\surd$ &  &  \\

    DisasterBench
    \citep{chen2026disasterbench}
    & $\surd$ &  &  &  &  &  & $\surd$ &  \\

    \midrule

    \textbf{This work}
    & $\surd$ & $\surd$ & $\surd$ & $\surd$
    & $\surd$ & $\surd$ & $\surd$ & $\surd$ \\

    \bottomrule
  \end{tabular*}

  \caption{
  Scope of related work along dimensions relevant to this study.
  The dimensions are:
  (a) typed workflow or tool-use outputs;
  (b) search compared with repeated sampling;
  (c) compute- or budget-controlled comparison;
  (d) candidate coverage separated from selected accuracy;
  (e) selector or verifier sensitivity;
  (f) attribution of performance differences among nominal inference
  procedures while controlling or auditing generation and
  post-generation pipeline factors;
  (g) localization of selected errors to tool or dependency decisions;
  and
  (h) recurrence of winning errors across procedures or independent
  runs.
  }
  \label{tab:related-positioning}
\end{table*}

Table~\ref{tab:related-positioning} compares research scope rather than
method quality or empirical performance.

\end{document}